\documentclass[review]{elsarticle}

\usepackage[margin=2.5cm]{geometry}
\usepackage{hyperref}
\usepackage{xurl}
\usepackage[T1]{fontenc}
\usepackage{setspace}
\usepackage{amsmath}
\usepackage{amssymb}
\usepackage{booktabs}
\usepackage{float}
\usepackage{enumerate}
\usepackage{enumitem}
\usepackage{xcolor}
\usepackage{appendix}
\usepackage{subcaption}

\hypersetup{ % set colours for hyperlinks
    colorlinks=true, 
    linkcolor=blue,
    filecolor=blue,      
    urlcolor=blue,
    citecolor=blue
}
\usepackage{cleveref}
\crefname{figure}{figure}{figures}
\Crefname{figure}{Figure}{Figures}
\Crefname{section}{Section}{Sections}
\Crefname{table}{Table}{Tables}

\usepackage{bbm}
\usepackage{multirow}
\usepackage[nolist,nohyperlinks]{acronym}
\usepackage{lscape}
\usepackage[numbers]{natbib} %not compatible with author-year citations, but needed for citep. 

\journal{The Ocular Surface}

\acrodef{CNN}{convolutional neural network} % Used >20 times
\acrodef{GAN}{generative adversarial network} % Keep abbreviation
\acrodef{SVM}{support vector machine} % Keep abbreviation
\acrodef{AI}{artificial intelligence} % Keep abbreviation
\acrodef{AUC}{area under the receiver operating characteristic curve}
\acrodef{TP}{true positive}
\acrodef{TN}{true negative}
\acrodef{FP}{false positive}
\acrodef{FN}{false negative}
\acrodef{DED}{Dry Eye Disease}
\acrodef{TBUT}{fluorescein tear break-up time}
\acrodef{AS-OCT}{anterior segment optical coherence tomography}
\acrodef{OCT}{optical coherence tomography}
\acrodef{OSDI}{Ocular Surface Disease Index}
\acrodef{IVCM}{in vivo confocal microscopy}
\acrodef{MGD}{meibomian gland dysfunction}
\acrodef{EU}{European Union}
\acrodef{FDA}{US Food and Drug Administration}
\begin{document}
\begin{frontmatter}

\title{Artificial Intelligence in Dry Eye Disease}

\author[1,3]{Andrea M.\ Stor{\aa}s\corref{c1}} %Main lead AI, writing, literature research, discussions
\cortext[c1]{Corresponding authors}
\ead{andrea@simula.no}
\author[1]{Inga Str{\"u}mke\corref{c1}} % Co lead AI, writing, literature research, discussions
\ead{inga@simula.no}
\author[1]{Michael A.\ Riegler} %Revision, discussions, medical
\author[2]{Jakob Grauslund} %Revision, discussions, medical
\author[1,3]{Hugo L.\ Hammer} %Revision 
\author[3]{Anis Yazidi} %Revision, references
\author[1,3]{P{\aa}l Halvorsen} % Revision, discussions AI
\author[6]{Kjell G.\ Gundersen} %revision discussions via Tor due to problems with overleaf
\author[3,4,5]{Tor P.\ Utheim} %Revision, discussions, medical
\author[6]{Catherine Jackson\corref{c1}} %main lead medical, writing, literature research, discussions 
\ead{catherinejoanjackson@gmail.com}

\address[1]{SimulaMet, Oslo, Norway}
\address[2]{Department of Ophthalmology, Odense University Hospital, Odense, Denmark}
\address[3]{Department of Computer Science, Oslo Metropolitan University, Norway}
\address[4]{Department of Medical Biochemistry, Oslo University Hospital, Norway}
\address[5]{Department of Ophthalmology, Oslo University Hospital, Norway}
\address[6]{Ifocus, Haugesund, Norway}

\begin{abstract}
Dry eye disease (DED) has a prevalence of between 5 and 50\%, depending on the diagnostic criteria used and population under study. However, it remains one of the most underdiagnosed and undertreated conditions in ophthalmology. Many tests used in the diagnosis of DED rely on an experienced observer for image interpretation, which may be considered subjective and result in variation in diagnosis. 
Since artificial intelligence (AI) systems are capable of advanced problem solving, use of such techniques could lead to more objective diagnosis. Although the term `AI' is commonly used, recent success in its applications to medicine is mainly due to advancements in the sub-field of machine learning, which has been used to automatically classify images and predict medical outcomes. Powerful machine learning techniques have been harnessed to understand nuances in patient data and medical images, aiming for consistent diagnosis and stratification of disease severity. This is the first literature review on the use of AI in DED. We provide a brief introduction to AI, report its current use in DED research and its potential for application in the clinic. Our review found that AI has been employed in a wide range of DED clinical tests and research applications, primarily for interpretation of interferometry, slit-lamp and meibography images. While initial results are promising, much work is still needed on model development, clinical testing and standardisation. 
\end{abstract}


\end{frontmatter}

\section{Introduction}
\ac{DED} is one of the most common eye diseases worldwide, with a prevalence of between 5 and 50\%, depending on the diagnostic criteria used and study population~\cite{stapleton2017tfos}. Yet, although symptoms stemming from \ac{DED} are reported as the most common reason to seek medical eye care~\cite{stapleton2017tfos}, it is considered one of the most underdiagnosed and undertreated conditions in ophthalmology~\cite{geerling2011international}. Symptoms of \ac{DED} include eye irritation, photophobia and fluctuating vision. The condition can be painful and might result in lasting damage to the cornea through irritation of the ocular surface. Epidemiological studies indicate that \ac{DED} is most prevalent in women~\cite{matossian2019dry} and increases with age~\cite{stapleton2017tfos}. However, the incidence of \ac{DED} is likely to increase in all age groups in coming years due to longer screen time and more prevalent use of contact lenses, which are both risk factors~\cite{nichols2005self}. Other risk factors include diabetes mellitus~\cite{zhang2016dry} and exposure to air-pollution~\cite{mandell2020impact}. \ac{DED} can have a substantial effect on the quality of life, and may impose significant direct and indirect public health costs as well as personal economic burden due to reduced work productivity.

\ac{DED} is divided into two subtypes defined by the underlying mechanism of the disease: (i) aqueous deficient \ac{DED}, where tear production from the lacrimal gland is insufficient and (ii) evaporative \ac{DED} (the most common form), which is typically caused by dysfunctional meibomian glands in the eyelids. Meibomian glands are responsible for supplying meibum, which is a concentrated substance that normally covers the surface of the cornea to form a protective superficial lipid layer that guards against evaporation of the underlying tear film. The ability to reliably distinguish between aqueous deficient and evaporative \ac{DED}, their respective severity levels and mixed aqueous/evaporative forms is important in deciding the ideal modality of treatment. A fast and accurate diagnosis relieves patient discomfort and also spares them  unnecessary expense and exposure to potential side effects associated with some treatments. A tailor made treatment plan can yield improved treatment response and maximize health provider efficiency.

The main clinical signs of \ac{DED} are decreased tear volume, more rapid break-up of the tear film (\ac{TBUT}) and microwounds of the ocular surface~\citep{willcox2017tfos}. In the healthy eye, the tear film naturally `breaks up' after ten seconds and the protective tear film is reformed with blinking. Available diagnostic tests often do not correlate with the severity of clinical symptoms reported by the patient. No single clinical test is considered definitive in the diagnosis of \ac{DED}\cite{stapleton2017tfos}. Therefore, multiple tests are typically used in combination and supplemented by information gathered on patient symptoms, recorded through questionnaires. These tests demand a significant amount of time and resources at the clinic. Tests for determining the physical parameters of tears include \ac{TBUT}, the Schirmer's test, tear osmolarity and tear meniscus height. Other useful tests in \ac{DED} diagnosis include ocular surface staining, corneal sensibility, interblink frequency, corneal surface topography, interferometry, aberrometry and imaging techniques such as meibography and \ac{IVCM}, as well as visual function tests. 

\ac{AI} was defined in 1955 as ``the science and engineering of making intelligent machines''~\cite{mccarthy2006AIproposal}, where intelligence is the ``ability to achieve goals in a wide range of environments''~\cite{legg2007intelligence}.
Within \ac{AI}, machine learning denotes a class of algorithms capable of learning from data rather than being programmed with explicit rules. \ac{AI}, and particularly machine learning, is increasingly becoming an integral part of health care systems. The sub-field of machine learning known as deep learning uses deep artificial neural networks, and has gained increased attention in recent years, especially for its image and text recognition abilities. In the field of ophthalmology, deep learning has so far mainly been used in the analysis of data from the retina to segment regions of interest in images, automate diagnosis and predict disease outcomes~\cite{schmidt2018artificial}. For instance, the combination of deep learning and \ac{OCT} technologies has allowed reliable detection of retinal diseases and improved diagnosis~\cite{deFauw2018clinically}. Machine learning also has potential for use in the diagnosis and treatment of anterior segment diseases, such as \ac{DED}. Many of the tests used for \ac{DED} diagnosis and follow-up rely on the experience of the observer for interpretation of images, which may be considered subjective~\cite{yedidya2007automatic}. \ac{AI} tools can be used to interpret images automatically and objectively, saving time and providing consistency in diagnosis. 

Several reviews have been published that discuss the application of \ac{AI} in eye disease, including screening for diabetic retinopathy~\cite{nielsen2019review}, detection of age-related macular degeneration~\cite{pead2019review} and diagnosis of retinopathy of prematurity~\cite{gensure2020review}. We are, however, not aware of any review on \ac{AI} in \ac{DED}. In this article, we therefore provide a critical review of the use of \ac{AI} systems developed within the field of \ac{DED}, discuss their current use and highlight future work.   
% ---------------------------------------------
\section{Artificial intelligence}\label{sec:AI}
% ---------------------------------------------
\ac{AI} is informational technology capable of performing activities that require intelligence. It has gained substantial popularity within the field of medicine due to its ability to solve ubiquitous medical problems, such as classification of skin cancer~\cite{esteva2017melanom}, prediction of hypoxemia during surgeries~\cite{lundberg2018hypoxemia} and identification of diabetic retinopathy~\cite{gulshan2016diabetic}. Machine learning is a sub-field of \ac{AI} encompassing algorithms capable of learning from data, without being explicitly programmed. All \ac{AI} systems used in the studies included in this review, fall within the class of machine learning. The process by which a machine learning algorithm learns from data is referred to as \textit{training}. The outcome of the training process is a machine learning \textit{model}, and the model's output is referred to as \textit{prediction}s. Different learning algorithms are categorised according to the type of data they use, and referred to as supervised, unsupervised and reinforcement learning. The latter is excluded from this review, as none of the studies use it, while the two former are introduced in this section. A complete overview of the algorithms encountered in the reviewed studies is provided in~\Cref{fig:overview_algorithms}, sorted according to the categories described below.
% ---------------------------------------------
\subsection{Supervised learning}
% ---------------------------------------------
Supervised learning denotes the learning process of an algorithm using labelled data, meaning data that contains the target value for each data instance, e.g., tear film lipid layer category. The learning process involves extracting patterns linking the input variables and the target outcome. The performance of the resulting model is evaluated by letting it predict on a previously unseen data set, and comparing the predictions to the true data labels. See~\Cref{sec:metrics} for a brief discussion of evaluation metrics. Supervised learning algorithms can perform regression and classification, where regression involves predicting a numerical value for a data instance, and classification involves assigning data instances to predefined categories. \Cref{fig:overview_algorithms} contains an overview of supervised learning algorithms encountered in the reviewed studies.
% ---------------------------------------------
\subsection{Unsupervised learning}
\begin{figure}[t!]
    \centering
    \includegraphics[width=\textwidth]{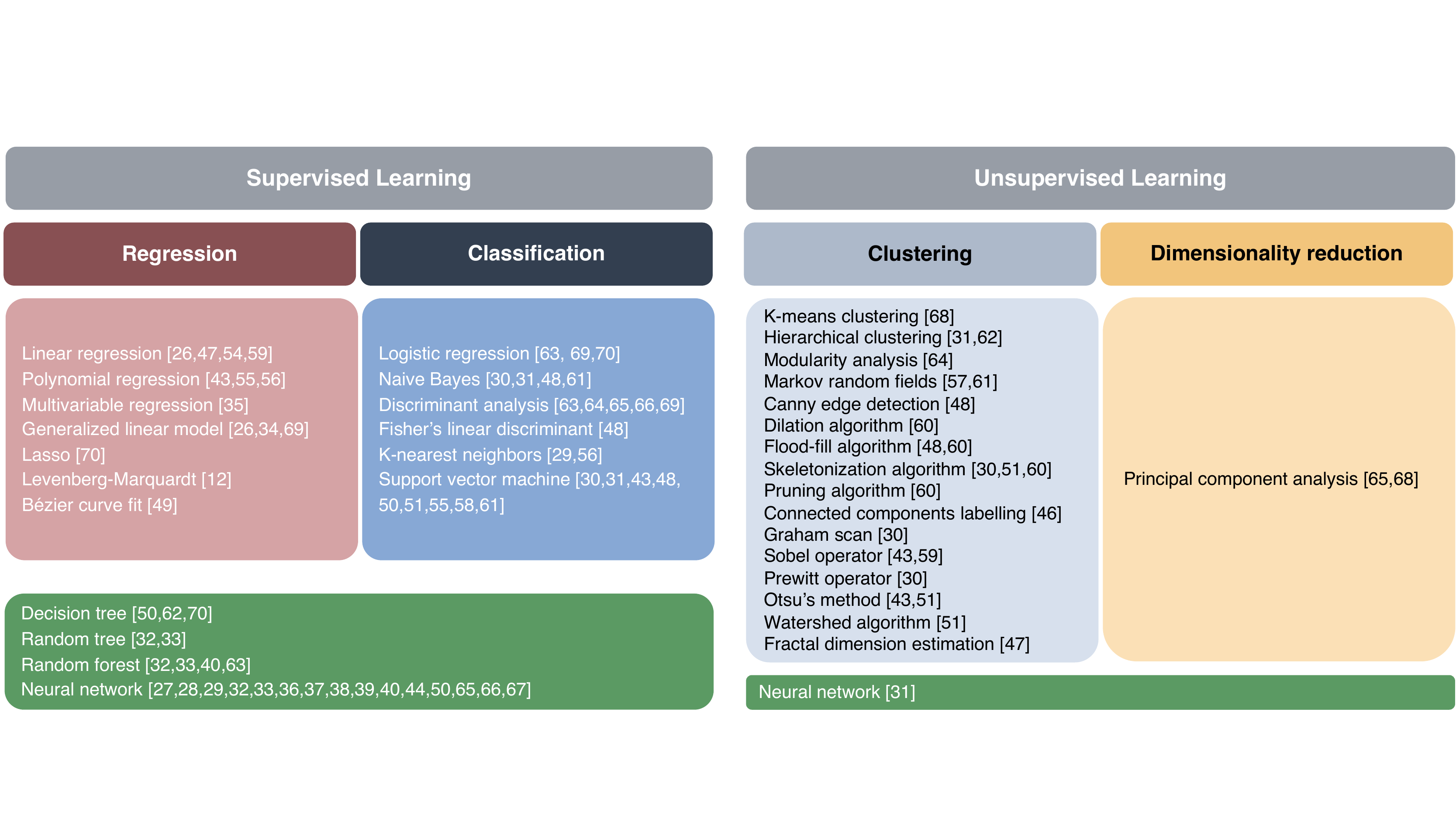}
    \caption{\label{fig:overview_algorithms}An overview of the machine learning algorithms used in the reviewed studies.}
\end{figure}

% ---------------------------------------------
Unsupervised learning denotes the training process of an algorithm using unlabelled data, i.e.\ data not containing target values. The task of the learning algorithm is to find patterns or data groupings by constructing a compact representation of the data. This type of machine learning is commonly used for grouping observations together, detecting relationships between input variables, and for dimensionality reduction. As unsupervised learning data contains no labels, a measure of model performance depends on considerations outside the data~\citep[see][chap. 14]{hastie2009Unsupervised}, e.g., how the task would have been solved by someone in the real world. For clustering algorithms, similarity or dissimilarity measures such as the distance between cluster points can be used to measure performance, but whether this is relevant depends on the task~\cite{palacio2019evaluation}.
Unsupervised algorithms encountered in the reviewed studies can be divided into those performing clustering and those used for dimensionality reduction, see~\Cref{fig:overview_algorithms} for an overview.
% ---------------------------------------------
\subsection{Artificial neural networks and deep learning}
% ---------------------------------------------
Artificial neural networks are loosely inspired by the neurological networks in the biological brain, and consist of artificial neurons organised in layers. How the layers are organised within the network is referred to as its \textit{architecture}. Artificial neural networks have one input layer, responsible for passing the data to the network, and one or more hidden layers. Networks with more than one hidden layer are called deep neural networks. The final layer is the output layer, providing the output of the entire network. Deep learning is a sub-field of machine learning involving training deep neural networks, which can be done both in a supervised and unsupervised manner. We encounter several deep architectures in the reviewed studies. The two more advanced types are \acp{CNN} and \acp{GAN}. \ac{CNN} denotes the commonly used architecture for image analysis and object detection problems, named for having so-called convolutional layers that act as filters identifying relevant features in images.
\acp{CNN} have gained popularity recently and all of the reviewed studies that apply \acp{CNN} were published in $2019$ or later. Advanced deep learning techniques will most likely replace the established image analysis methods. This trend has been observed within other medical fields such as gastrointestinal diseases and radiology~\cite{le2020application,thrall2018artificial}.
A \ac{GAN} is a combination of two neural networks: A generator and a discriminator competing against each other. The goal of the generator is to produce fake data similar to a set of real data. The discriminator receives both real data and the fake data from the generator, and its goal is to discriminate the two. \acp{GAN} can be used i.a.\ to generate synthetic medical data, alleviating privacy concerns~\citep{thambawita2021id}.

% ---------------------------------------------
\subsection{Workflow for model development and validation}
% ---------------------------------------------
The data used for developing machine learning models is ideally divided into three independent parts: A training set, a validation set and a test set. The training set is used to tune the model, the validation set to evaluate performance during training, and the test set to evaluate the final model. A more advanced form of training and validation, is $k$-fold cross-validation. Here, the data is split into $k$ parts, of which one part is set aside for validation, while the model is trained on the remaining data. This is repeated $k$ times, and each time a different part of the data is used for validation. The model performance can be calculated as the average performance for the $k$ different models~\citep[see][chap. 7]{hastie2009Unsupervised}. It is considered good practice to not use the test data during model development and vice versa, the model should not be tuned further once it has been evaluated on the test data~\citep[see][chap.7]{hastie2009Unsupervised}. In cases of class imbalance, i.e., unequal number of instances from the different classes, there is a risk of developing a model that favors the prevalent class. If the data is stratified for training and testing, this might not be captured during testing. Class imbalance is common in medical data sets, as there are for instance usually more healthy than ill people in the population~\cite{gianfrancesco2018classImbalance}. Whether to choose a class distribution that represents the population, a balanced or some other distribution depends on the objective. Various performance scores should regardless always be used to provide a full picture of the model's performance.
% ---------------------------------------------
\subsection{Performance scores}\label{sec:metrics}
% ---------------------------------------------
In order to assess how well a machine learning model performs, its performance can be assigned a score. In supervised learning, this is based on the model's output compared to the desired output. Here, we introduce scores used most frequently in the reviewed studies. Their definitions as well as the remaining scores used are provided in~\Cref{sec:metrics_appendix}.
A commonly used performance score in classification is \textit{accuracy}, eq.~\eqref{eq:accuracy}, which denotes the proportion of correctly predicted instances. Its use is inappropriate in cases of strong class imbalance, as it can reach high values if the model always predicts the prevalent class. The \textit{sensitivity}, also known as recall, eq.~\eqref{eq:recall}, denotes the true positive rate. If the goal is to detect all positive instances, a high sensitivity indicates success. The \textit{precision}, eq.~\eqref{eq:precision}, denotes the positive predictive value. The \textit{specificity}, eq.~\eqref{eq:specificity}, denotes the true negative rate, and is the negative class version of the sensitivity. The \textit{F1 score}, eq.~\eqref{eq:f1_score}, is the harmonic mean between the sensitivity and the precision. It is not symmetric between the classes, meaning it is dependent on which class is defined as positive.

Image segmentation involves partitioning the pixels in an image into segments~\cite{geron2019hands}. This can for example be used to place all pixels representing the pupil into the same segment while pixels representing the iris are placed in another segment. The identified segments can then be compared to manual annotations. Performance scores used include the \textit{Average Pompeiu-Hausdorff distance}, \eqref{eq:aph}, the \textit{Jaccard index} and the \textit{support}, all described in~\Cref{sec:metrics_appendix}. 
% ---------------------------------------------

\subsection{AI regulation}
Approved \ac{AI} devices will be a major part of the medical service landscape in the future.  Currently, many countries are actively working on releasing \ac{AI} regulations for healthcare, including the \ac{EU}, the United States, China, South Korea and Japan. 
On 21 April 2021, the \ac{EU} released a proposal for a regulatory framework for \ac{AI}~\cite{AI_regulation_EU}. The \ac{FDA} is also working on \ac{AI} legislation for healthcare~\cite{AI_regulation_FDA}.

In the framework proposed by the \ac{EU}, \ac{AI} systems are divided into the four categories low risk, minimal risk, high risk and unacceptable risk~\cite{AI_regulation_EU}. \ac{AI} systems that fall into the high risk category are expected to be subject to strict  requirements, including data governance, technical documentation, transparency and provision of information to users, human oversight, robustness and cyber security, and accuracy. It is highly likely that medical devices using \ac{AI} will end up in the high risk category.
Looking at the legislation proposals~\cite{AI_regulation_EU,AI_regulation_FDA} from an \ac{AI} research perspective, it is clear that explainable \ac{AI}, transparency, uncertainty assessment, robustness against adversarial attacks, high quality of data sets, proper performance assessment, continuous post-deployment monitoring, human oversight and interaction between \ac{AI} systems and humans, will be major research topics for the development of \ac{AI} in healthcare.

\section{Methods}\label{sec:methods}

\subsection{Search methods}
\begin{figure}[tb!]
    \centering
    \includegraphics[width=0.7\textwidth]{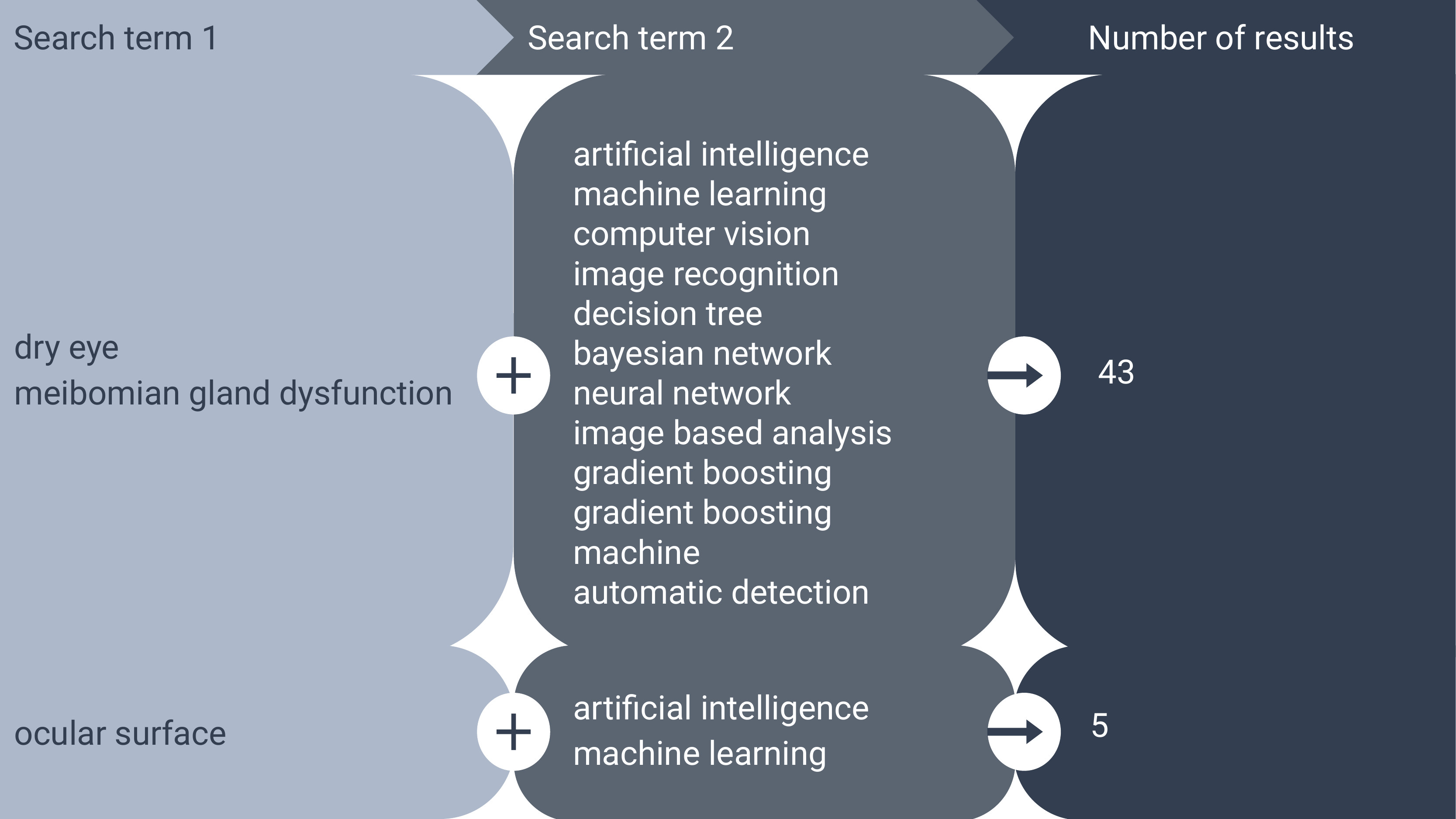}
    \caption{\label{tab:search_terms}Search term combinations used in the literature search. Three of the studies found in the searches including ``ocular surface'' were also found among the studies in the searches including ``dry eye''.}
\end{figure}

% ---------------------------------------------
A systematic literature search was performed in PubMed and Embase in the period between March $20$ and May $21$, $2021$. The goal was to retrieve as many studies as possible applying machine learning to \ac{DED} related data. The following keywords were used: All combinations of ``dry eye'' and ``meibomian gland dysfunction'' with ``artificial intelligence'', ``machine learning'', ``computer vision'', ``image recognition'', ``bayesian network'', ``decision tree'', ``neural network'', ``image based analysis'', ``gradient boosting'', ``gradient boosting machine'' and ``automatic detection''. In addition, searches for ``ocular surface'' combined with both ``artificial intelligence'' and ``machine learning'' were made. See also an overview of the search terms and combinations in~\Cref{tab:search_terms}. No time period limitations were applied for any of the searches. 
% ---------------------------------------------
% ---------------------------------------------
\subsection{Selection criteria}
% ---------------------------------------------
The studies to include in the review had to be available in English in full-text. Studies not investigating the medical aspects of \ac{DED} were excluded (e.g., other ocular diseases and cost analyses of \ac{DED}). Moreover, the studies had to describe the use of a machine learning model in order to be considered. Reviews were not considered. The studies were selected in a three-step process. One review author screened the titles on the basis of the inclusion criteria. The full-texts were then retrieved and studied for relevance.   
The search gave $640$ studies in total, of which $111$ were regarded as relevant according to the selection criteria. After removing duplicates, $45$ studies were left. The three-step process is shown in~\Cref{fig:flowchart}.
% ---------------------------------------------
\section{Artificial intelligence in dry eye disease}\label{sec:studies}
% ---------------------------------------------
\subsection{Summary of the studies}\label{sec:results}

\begin{figure}[!tb]
    \centering
    \begin{subfigure}{.49\linewidth}{
    \includegraphics[width=\textwidth]{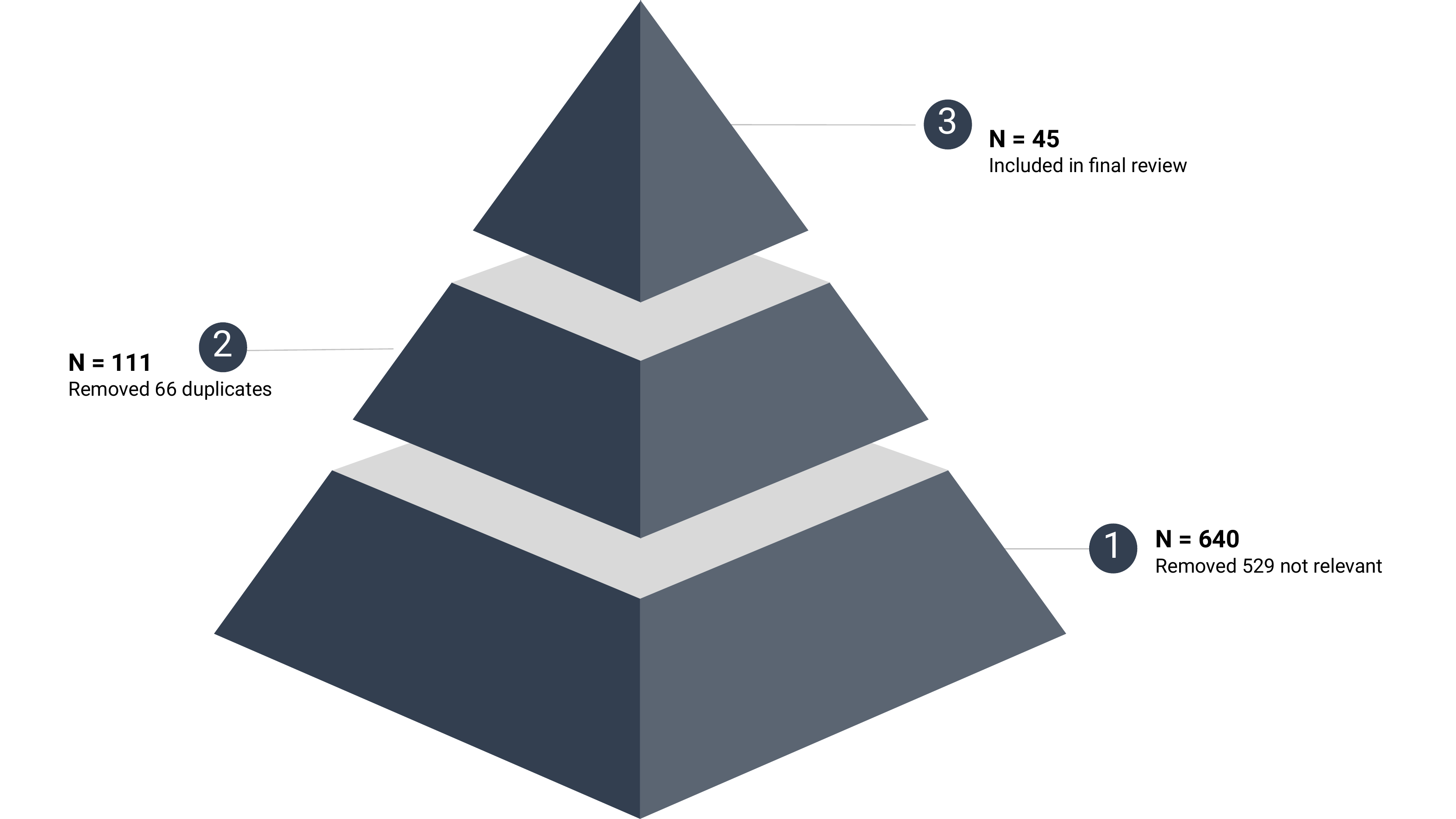}
    \caption{\label{fig:flowchart}}}
    \end{subfigure}
    \begin{subfigure}{.49\linewidth}{
    \includegraphics[width=\textwidth]{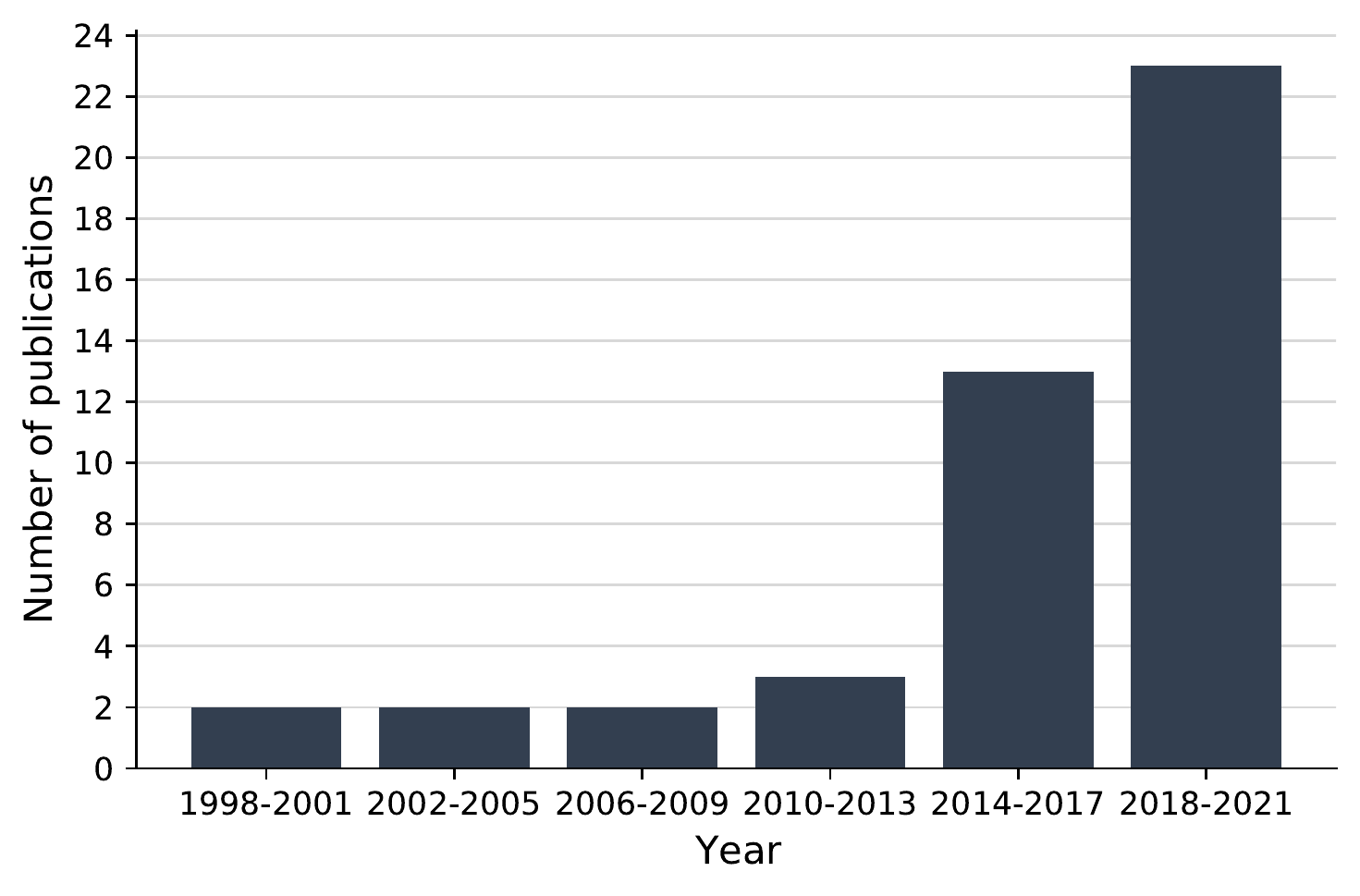}
    \caption{\label{fig:timechart}}}
    \end{subfigure}
    \caption{(\protect\subref{fig:flowchart})
    Illustration of the three steps in the study selection process and number of studies (\textbf{N}) included in each step, and 
    (\protect\subref{fig:timechart})
    the number of studies published over time, counting the studies included in this review.}
\end{figure}

Most studies were published in recent years, especially after 2014, see~\Cref{fig:timechart}. An overview of the studies is provided in~\Cref{tab:clinicaltable_1,tab:clinicaltable_2,tab:biochemicaltable,tab:populationtable} for the clinical, biochemical and demographical studies, respectively. Information on the data used in each study is shown in~\Cref{tab:datatable}. We grouped studies according to the type of clinical test or type of study: \ac{TBUT}, interferometry and slit-lamp images, \ac{IVCM}, meibography, tear osmolarity, proteomics analysis, \ac{OCT}, population surveys and other clinical tests. We found most studies employed machine learning for interpretation of interferometry, slit-lamp and meibography images. 

\begin{landscape}
\begin{table}
%   \scriptsize
    \tiny
    \caption{\label{tab:clinicaltable_1}Overview of the reviewed studies using clinical investigations, part 1 of 2.
    }
    \centering
    \begin{tabular}{p{0.1\textwidth}p{0.1\textwidth}
    p{0.05\textwidth}p{0.2\textwidth}p{0.1\textwidth}
    p{0.2\textwidth}p{0.4\textwidth}}  
    \toprule 
         Study & Objective & N & Clinical Tests & Type of Data & Type of Algorithm & Performance Score(s)  \\
         \midrule
         %\textbf{Clinical Investigations} \\
         Aggarwal S et al. ($2021$)~\cite{aggarwal2021immunecell} & DED mechanism, effect of therapy & $199$ & Subjective symptoms, Schirmer´s test with anasthesia, TBUT, vital staining of cornea and conjunctiva, laser IVCM images, subbasal layer of cornea: DC density and morphology & Images of cornea & GLM, MLR & GLM: p-values < $0.05$ for DC density and number of DCs, MLR: p-values < $0.05$ between DC density and CFS, number of DCs and CFS, DC size and CFS, DC density and conjunctival staining, number of DCs and TBUT, corresponding $\beta$-coefficients = $0.20$, $-0.23$, $0.36$, $0.24$ and $-0.18$ \\
         Deng X et al. ($2021$)~\cite{DENG2021TearMeniscus} & Estimate tear meniscus height & $217$ & Oculus Keratograph & Tear meniscus images & CNN (U-net) & Accuracy = $82.5$\%, sensitivity = $0.899$, precision = $0.911$, F1 score = $0.901$ \\
         
         Elsawy A et al. ($2021$)~\cite{ELSAWY2021252} & Diagnose DED & $547$ & AS-OCT & Ocular surface images & Pretrained CNN (VGG19) & AUCROC = $0.99$ (model $1$) and $0.98$ (model $2$), AUCPRC = $0.96$ (model $1$) and $0.94$ (model $2$), F1 score = $0.90$ (model $1$) and $0.86$ (model $2$)* \\
         Khan ZK et al. ($2021$)\cite{Khan2021image} & Detect MGD & $112$ & Meibomian gland 3D IR-images, lower and upper eyelid & Meibomian gland images & GAN & F1 score = $0.825$, P-HD = $4.611$, aggregated JI = $0.664$, r = $0.962$ (clincian1) and $0.968$ (clinician2), p-values < $0.001$, mean difference = $0.96$ (clincian1) and $0.95$ (clincian2) \\
         Xiao P et al.($2021$)~\cite{xiao2021Meibo} & Detect MGD & $15$ (images) & Oculus Keratograph & IR meibography images & Prewitt operator, Graham scan algorithm, fragmentation algorithm and SA (used sequentially) & Gland area: KI = $0.94$, FPR = $6.02$\%, FNR = $6.43$\%. Gland segmentation: KI = $0.87$, FPR = $4.35$\%, FNR = $18.61$\%*  \\
         Yeh C-H et al. ($2021$)~\cite{yeh2021Meibo} & Detect MGD & $706$ (images) & Oculus Keratograph & IR meibography images & Nonparametric instance discrimination, pretrained CNN (ImageNet), hierarchical clustering & : Accuracy: meiboscore grading = $80.9$\%, 2-class classification = $85.2$\%, 3-class classification = $81.3$\%, 4-class classification = $80.8$\%*  \\
         da Cruz LB et al. ($2020$)\cite{dacruz2020interferometer} & Classify tear film patterns & $106$ (images) & Doane interferometer & Tear film lipid layer images & SVM, RF, RT, Naive Bayes, DNN, simple NN & RF: accuracy = $97.54$\%, SD = $0.51$\%, F1 score = $0.97$, KI = $0.96$, AUCROC = $0.99$**  \\
         da Cruz LB et al. ($2020$)~\cite{dacruz2020ripleysk} & Classify tear film patterns & $106$ (images) & Doane interferometer & Tear film lipid layer images & SVM, RF, RT, Naive Bayes, DNN, simple NN  & RF: accuracy = $99.622$\%, SD = $0.843$\%, F1 score = $0.996$, KI = $0.995$, AUCROC = $0.999$*** \\
         Fu P-I et al. ($2020$)~\cite{Fu2020LLT} & Compare $2$ methods & $28$ & Oculus Keratograph & Tear film lipid layer images (with and without preprocessing) & GLM & $\beta$-coefficients = $0.6$, $10$ \\ 
         Fujimoto K et al. ($2020$)~\cite{fujimoto2020comparison} & Compare $2$ methods & $195$ & Pentacam vs AS-OCT  & CCT, TCT, thinnest point of cornea & Multivariable regression & Severe DED: $\beta$-coefficients = $7.029$ (CCT) and $6.958$ (TCT), p-values = $0.002$ (CCT) and $0.049$ (TCT), $95$\% CI = $2.528-11.530$ (CCT) and $0.037-13.879$ (TCT) \\
         Maruoka S et al. ($2020$)~\cite{Maruoka2020Meibo} & Detect MGD & $221$ & IVCM & Meibomian gland images & Combinations of $9$ CNNs & Single CNN: AUROC = $0.966$, sensitivity = $0.942$, specificity = $0.821$, ensemble CNNs: AUROC = $0.981$, sensitivity = $0.921$, specificity = $0.988$ \\
         Prabhu SM et al. ($2020$)~\cite{Prabhu2020Meibo} & Quantify and detect MGD & $400$ (images) & Oculus Keratograph, digital camera & CNN (U-net) & $p$-values > $0.005$ between model output and clinical experts \\
         Stegmann H et al. ($2020$)~\cite{Stegmann2020TearMeniscus} & Detect tear meniscus in images & $10$ & Optical coherence tomography & Tear meniscus images & $2$ CNNs & Meniscus localization: JI = $0.7885$, sensitivity = $0.9999$, meniscus segmentation best CNN: accuracy = $0.9995$, sensitivity = $0.9636$ , specificity = $0.9998$, JI = $0.9324$, F1 score = $0.9644$, support = $0.0071$*, *** \\
         Wei S et al. ($2020$)~\cite{wei2020therapeutic} & DED mechanism, effect of therapy & $53$ & Corneal IVCM with anesthesia & Images of cornea & Pretrained CNN (U-net) & AUROC = $0.96$,  sensitivity = $96$\% \\
         Giannaccare G et al. ($2019$)~\cite{Giannaccare2019SubbasalNerve} & Subbasal nerve plexus characteristics for diagnosing DED & $69$ & IVCM & Images of subbasal nerve plexus & Earlier developed method involving RF and NN~\cite{Chen2017ACCMed, Dabbah2011NerveFiber} & Nan \\
         \bottomrule
         \multicolumn{7}{p{\linewidth}}{Abbreviations: N = number of subjects; DED = dry eye disease; IVCM = in vivo confocal microscopy; DC = dendritic cell; GLM = generalized linear model; MLR = multiple linear regression; CFS = corneal fluorescein score;  AS-OCT = anterior segment optical coherence tomography; CNN = convolutional neural network; AUROC = area under reciever operating characteristic curve; AUPRC = area under precision-recall curve; MGD = meibomian gland dysfunction; GAN = generative adversarial network; P-HD = average Pompeiu-Hausdorff distance; JI = Jaccard index; KI = Kappa index; CTRL = healthy; FPR = false positive rate; FNR = false negative rate; SVM = support vector machine; RF = random forest; RT = random tree; DNN = deep neural network; SD = standard deviation; CCT = central corneal thickness; TCT = thinnest corneal thickness; r = Pearson's correlation coefficient;Nan = not available; NN = neural network; RMSE = root mean squared error; CI = confidence interval;  TBUT = fluorescein tear break-up time; PA = pruning algorithm; SA = skeletonization algorithm; FFA = Flood-fill algorithm; * = standard deviations not included in table; ** = $95$\% confidence intervals not included in table; *** = metrics are calculated as the average of $5$ repetitions;**** = metrics are calculated as the average of $10$ repetitions; ***** = metrics are calculated as the average from $10$-fold crossvalidation; $\ominus$ = metrics are calculated as the average from $6$-fold crossvalidation $\oplus$ = metrics are calculated as the average of $100$ models)} \\
    \end{tabular}
\end{table}
\end{landscape}
\begin{landscape}
\begin{table}
%   \scriptsize
    \tiny
    \caption{\label{tab:clinicaltable_2}Overview of the reviewed studies using clinical investigations, part 2 of 2.
    }
    \centering
    \begin{tabular}{p{0.1\textwidth}p{0.1\textwidth}
    p{0.05\textwidth}p{0.2\textwidth}p{0.1\textwidth}
    p{0.2\textwidth}p{0.4\textwidth}} 
    \toprule 
         Study & Objective & N & Clinical Tests & Type of Data & Type of Algorithm & Performance Score(s)  \\
         \midrule
         %\textbf{Clinical Investigations} \\
        Llorens-Quintana C et al. ($2019$)~\cite{Llorens-QuintanaClara2019ANAA} & Evaluate meibomian gland atrophy & $149$ & Oculus Keratograph & Meibography images & Sobel operator, polynomial function, fragmentation algorithm, Otsu's method (used sequentially) & $p$-values < $0.05$ between automatic method and clinicians \\
        Wang J et al. ($2019$)~\cite{Wang2019Meibo} & Evaluate meibomian gland atrophy & $706$ (images) & Oculus Keratograph & Meibography images & $4$ CNNs & Meiboscore grading: accuracy = $95.6$\%, eyelid detection: accuracy = $97.6$\%, JI = $0.955$, atrophy detection: accuracy = $95.4$\%, JI = $0.667$, RMSE = $0.067$ (average across $4$ meiboscores) \\
         Yabusaki K ($2019$)~\cite{yabusaki2019diagnose} & Diagnose DED & $138$ (images) & Tear interferometer & Tear film lipid layer images & SVM & KI = $0.820$, CTRL: F1 score = $0.845$, SD = $0.067$, aqueous-deficient DED: F1 score = $0.981$, SD = $0.023$, evaporative DED: F1 score = $0.815$, SD = $0.095$**** \\
         Yang J et al. ($2019$)~\cite{yang2019meniscus} & Estimate tear meniscus height for DED & $69$ & Slit-lamp images with fluorescence staining & Ocular surface images & Connected component labelling & Mean: p-value < $0.01$ (x$16$ and x$40$ magnification), r = $0.626$ (x$16$) and  $0.711$ (x$40$), max: p-value < $0.001$ (x$16$ and x$40$), r = $0.645$ (x$16$) and $0.847$ (x$40$) \\
         Szyperski PD ($2018$)~\cite{Szyperski2018fractal} & Diagnose DED & $110$ & Interferometry & Videos from lateral shearing interferometry & $4$ different fractal dimension estimators, linear regression & Best estimator: AUROC = $0.786$ \\
         Hwang H et al. ($2017$)~\cite{hwang2017image} & Estimate tear film lipid layer thickness & $34$ & Lipiscanner $1.0$, slit-lamp microscope & Tear film lipid layer videos & Flood-fill algorithm, Canny edge detection & p-value < $0.01$ between all MGD groups \\
         Koprowski R et al. ($2017$)~\cite{Koprowski2017Meibo} & Detect MGD & $57$ & Oculus Keratograph & Meibography images & Riesz pyramid (?), Bezier curve (used sequentially) & Accuracy = $99.08$\%, sensitivity = $1$, specificity = $0.98$\\
         Peteiro-Barral D et al. ($2017$)~\cite{peteiro2017evaluation} & Classify tear film patterns & $105$ (images) & Tearscope plus images & Tear film lipid layer images & SVM, Decision tree, Naive Bayes, simple NN, Fisher´s linear discriminant & NN: accuracy = $96$\%, sensitivity = $92$\%, specificity = $97$\%, precision = $92$\%, F1 score = $0.93$, AUCROC = $0.95$ \\ 
         Koprowski et al. ($2016$)~\cite{Koprowski20161Meibo} & Detect MGD & $86$ & Oculus Keratograph & Meibography images & Otsu's method, SA, watershed algorithm (used sequentially) & Sensitivity = $0.993$, specificity = $0.975$ \\
         Remeseiro B et al. ($2016$)~\cite{remeseiro2016iDEAS} & Classify tear film patterns & $128$ (images) & Tearscope-plus images & Tear film lipid layer images & SVM & accuracy = $96.09$\%, precision = $92.00$\%, sensitivity = $89.66$\%, specificity = $97.98$\%, F1 score = $91.23$\%, processing time = $0.07$ s\\
         Remeseiro B et al. ($2016$)~\cite{remeseiro2016CASDES} & Classify tear film patterns & $50$ (images) & Tearscope-plus images & Tear film lipid layer images & SVM & accuracy = $90.89$\%, sensitivity = $83.54$\%, precision = $97.95$\%, specificity = $86.75$\% \\
         Kanellopoulos AJ et al. ($2014$)~\cite{kanellopoulos2014invivo} & Diagnose DED & $70$ & Fourier-domain AS-OCT system: corneal and corneal epithelial thickness maps & Corneal examination & Linear regression (correlation between DED and thickness) & Nan \\
         Ramos L et al. ($2014$)~\cite{ramos2014automatic} & Estimate TBUT & $18$ (videos) & Videos from TBUT (slit-lamp) & TBUT videos & Polynomial function & specificity = $89$\% (parameter b) and $82$\% (parameter e), specificity = $84$\% and $80$\% \\
         Ramos L et al. ($2014$)~\cite{Ramos2014CCLRU} & Estimate TBUT & $18$ (videos) & Videos from TBUT (slit-lamp) & TBUT videos & Polynomial function & accuracy = "more than $90$\%" \\
         Remeseiro et al. ($2014$)~\cite{remeseiro2014tearfilm} & Classify tear film patterns & $511$ (images) & Tearscope-plus images & Tear film lipid layer images & Markov random field, SVM (used sequentially) & accuracy = $97.14$\%, accuracy (noisy data) = $92.61\%$*****\\ 
         García-Resúa C et al. ($2013$)~\cite{garcia2013new} & Classify tear film patterns & $105$ & Tearscope-plus images & Tear film lipid layer images & K-nearest neighbors & Cram{\'e}r's V = $0.9$, r = $0.94$, p-value < $0.001$, accuracy = $86.2$\% $\ominus$\\
         Rodriquez JD ($2013$)~\cite{Rodriguez2013Redness} & Evaluate ocular redness & $26$ & Slit-lamp, digital camera & Images of conjunctiva & Sobel operator, MLR (used sequentially) & Accuracy = $100$\%, r = $0.76$, concordance correlation = $0.76$ (compared to $5$ investigators)*\\
         Koh YW et al. ($2012$)~\cite{koh2012detection} & Detect MGD & $55$ & Slit-lamp biomicroscope, upper eye lid & IR meibography images & PA, SA, FFA, SVM (used sequentially) & specificity = $96.1$\%, SD = $0.4$\%, sensitivity = $97.9$\%, SD = $0.6$\% $\oplus$\\
         Yedidya T et al. ($2009$)~\cite{yedidya2009enforcing} & Estimate TBUT & $22$ (videos) & Video from TBUT & TBUT videos & Markov random field  & average difference in TBUT = $2.34$s \\
         Yedidya T et al. ($2007$)~\cite{yedidya2007automatic} & Detect dry areas & $8$ (videos) & Video from TBUT & TBUT videos & Levenberg-Marquardt & accuracy = $91$\% ($84-96$\%), SD = $4$\%\\
         Mathers WD et al. ($2004$)~\cite{mathers2004cluster} & Investigate DED & $513$ & Schirmer´s test, meibomian gland drop-out, lipid viscosity and volume, tear evaporation & Clinical test results & Hierarchical clustering, decision tree & Nan\\
         \bottomrule
         \multicolumn{7}{p{\linewidth}}{Abbreviations: N = number of subjects; DED = dry eye disease; IVCM = in vivo confocal microscopy; DC = dendritic cell; GLM = generalized linear model; MLR = multiple linear regression; CFS = corneal fluorescein score;  AS-OCT = anterior segment optical coherence tomography; CNN = convolutional neural network; AUROC = area under reciever operating characteristic curve; AUPRC = area under precision-recall curve; MGD = meibomian gland dysfunction; GAN = generative adversarial network; P-HD = average Pompeiu-Hausdorff distance; JI = Jaccard index; KI = Kappa index; CTRL = healthy; FPR = false positive rate; FNR = false negative rate; SVM = support vector machine; RF = random forest; RT = random tree; DNN = deep neural network; SD = standard deviation; CCT = central corneal thickness; TCT = thinnest corneal thickness; r = Pearson's correlation coefficient;Nan = not available; NN = neural network; RMSE = root mean squared error; CI = confidence interval;  TBUT = fluorescein tear break-up time; PA = pruning algorithm; SA = skeletonization algorithm; FFA = Flood-fill algorithm; * = standard deviations not included in table; ** = $95$\% confidence intervals not included in table; *** = metrics are calculated as the average of $5$ repetitions;**** = metrics are calculated as the average of $10$ repetitions; ***** = metrics are calculated as the average from $10$-fold crossvalidation; $\ominus$ = metrics are calculated as the average from $6$-fold crossvalidation $\oplus$ = metrics are calculated as the average of $100$ models)} \\
    \end{tabular}
\end{table}
\end{landscape}
\begin{landscape}
\begin{table}
%   \scriptsize
    \tiny
    \caption{\label{tab:biochemicaltable}Overview of the reviewed studies using biochemical investigations.
    }
    \centering
    \begin{tabular}{p{0.1\textwidth}p{0.1\textwidth}
    p{0.05\textwidth}p{0.2\textwidth}p{0.1\textwidth}
    p{0.2\textwidth}p{0.4\textwidth}} 
    %\begin{tabular}{p{2.3cm}p{2.7cm}p{1cm}p{3cm}p{2.5cm}p{3cm}p{6.3cm}} 
    \toprule 
         Study & Objective & N & Clinical Tests & Type of Data & Type of Algorithm & Performance Score(s)  \\
         \midrule
         %\textbf{Biochemical Investigations} \\
         
         Cartes C et al. ($2019$)~\cite{cartes2019dry} & Diagnose DED & $40$ & Tear-Lab Osmometer & Tear osmolarity measurements & LR, Naive Bayes, SVM, RF & LR: accuracy = $85$\%\\
         Jung JH et al. ($2017$)~\cite{jung2017proteomic} & Detect protein patterns in DED & $10$ & Pooled tear and lacrimal fluid, analysed with LC-MS, trypsin digestion, RP-LC fractionation & Proteins in tears and lacrimal fluid & "Network model" based on betweenness centrality & Nan\\
         Gonzalez N ($2014$)~\cite{Gonzalez2014Protein} & Diagnose DED & $93$ & Peptide/protein analysis: gel electrophoresis (SDS-PAGE) & Peptides and proteins in tears & Discriminant analysis, PCA, NN & Accuracy = $89.3$\%, CTRL: sensitivity = $0.99$, specificity = $0.96$, MGD: sensitivity = $0.85$, specificity = $0.96$, aqueous-deficient DED: sensitivity = $0.83$, specificity = $0.93$* \\
         Grus FH et al. ($2005$)~\cite{grus2005seldi} & Diagnose DED & $159$ & Schirmer´s test with anesthesia, tears analysed by LC-MS & Proteins in tears & Discriminant analysis, DNN (used sequentially) & AUROC = $0.93$, sensitivity and specificity = ``approx. $90$\% each''\\
         Grus FH et al. ($1999$)~\cite{grus1999analysis} & Diagnose DED & $60$ & Protein analysis: gel electrophoresis (SDS-PAGE) & Proteins in tears  & DNN, discriminant analysis & DNN: accuracy = $89$\%, discriminant analysis: accuracy = $71$\%\\
         Grus FH et al. ($1998$)~\cite{grus1998clustering} & Diagnose DED & $119$ & Protein analysis: gel electrophoresis (SDS-PAGE) & Proteins in tears & Principal component analysis, K-means clustering (used sequentially), discriminant analysis & K-means: accuracy = $71$\% (DED vs CTRL) and $42$\% (DED, diabetes-DED, CTRL), discriminant analysis: accuracy = $72$\% (DED vs CTRL) and $43$\% (DED, diabetes-DED, CTRL) \\
         \bottomrule
         \multicolumn{7}{p{\linewidth}}{Abbreviations: N = number of subjects; DED = dry eye disease; LR = logistic regression; SVM = support vector machine; RF = random forest; AUROC = area under reciever operating characteristic curve; MGD = meibomian gland dysfunction; CTRL = healthy; DNN = deep neural network; Nan = not available; NN = neural network; LC-MS = liquid chromotography mass spectometry; RP-LC = reverse-phase liquid chromatography; SDS-PAGE = sodium dodecyl sulphate-polyacrylamide gel electrophoresis; OSDI = ocular surface disease index; * = metrics are calculated as the average of $10$ repetitions} \\
    \end{tabular}
\end{table}

\begin{table}
%   \scriptsize
    \tiny
    \caption{\label{tab:populationtable}Overview of the reviewed studies using demographical investigations.
    }
    \centering
   \begin{tabular}{p{0.1\textwidth}p{0.1\textwidth}
    p{0.05\textwidth}p{0.2\textwidth}p{0.1\textwidth}
    p{0.2\textwidth}p{0.4\textwidth}}  
    %\begin{tabular}{p{2.3cm}p{2.7cm}p{1cm}p{3cm}p{2.5cm}p{3cm}p{6.3cm}}  % Backup
    \toprule 
         Study & Objective & N & Clinical Tests & Type of Data & Type of Algorithm & Performance Score(s)  \\
         \midrule
         %\textbf{Demographical Investigations} \\
         
         Choi HR et al. ($2020$)~\cite{choi2020association} & Investigate DED and dyslipidemia association & $2272$ & OSDI score, health examination, questionnaire & Population studies, Korea & GLM, LR & Nan\\
         Nam SM et al. ($2020$)~\cite{nam2020explanatory} & Detect risk factors for DED & $4391$ & Health examination, health survey, nutrition survey & National health survey, Korea &  Decision tree, Lasso, LR (used sequentially) & AUROC = $0.70$,  $95$\% CI = $0.61 - 0.78$, specificity = $68$\%, sensitivity = $66$\%\\
         Kaido M et al. ($2015$)~\cite{Kaido2015computer} & Diagnose DED & $369$ & Blink frequency, visual maintenance ratio, questionnaire & Functional VA measurement and questionnaire, Japanese visual display terminal workers & Discriminant analysis & sensitivity = $93.1$\%, specificity = $43.7$\%, precision = $83.8$\%, NPV = $80.8$\%\\
         \bottomrule
         \multicolumn{7}{p{\linewidth}}{Abbreviations: N = number of subjects; DED = dry eye disease; GLM = generalized linear model; AUROC = area under reciever operating characteristic curve; Nan = not available; CI = confidence interval;  LR = logistic regression; OSDI = ocular surface disease index; VA = visual acuity; NPV = negative predictive value} \\
    \end{tabular}
\end{table}
\end{landscape}
\begin{landscape}
\begin{table}
    \caption{\label{tab:datatable}Overview of the data applied for the analyses. %Abbreviations: CV = crossvalidation; * = For multivariate analysis model, but the number of samples was not mentioned
    }
    \centering
    \tiny
    \begin{tabular}{lllll} 
    %\begin{tabular}{llp{3cm}p{3cm}l} 
    %\begin{tabular}{p{0.2\textwidth}p{0.2\textwidth}
    %p{0.1\textwidth}p{0.3\textwidth}p{0.2\textwidth}
    %p{0.2\textwidth}p{0.4\textwidth}} 
    \toprule 
         Study & Type of Input Data & Training Dataset & Testing Dataset & Reference Standard  \\
         \midrule
         \textbf{Clinical Investigations}   \\
         Aggarwal S et al. ($2021$)~\cite{aggarwal2021immunecell} & Tabular & $349$ & Nan & Nan (clinical test results, subjective report)  \\
         Deng X ($2021$)~\cite{DENG2021TearMeniscus} & Images & $253$ (images) & $232$ (images) & Senior clinician\\
         Elsawy A et al. ($2021$)~\cite{ELSAWY2021252} & Images & $29172$ (train), $7293$ (val) & $23760$ & Certified cornea specialist  \\
         Khan ZK et al. ($2021$)\cite{Khan2021image} & Images & $90$ & $22$ & Clinician  \\
         Xiao P et al. ($2021$)~\cite{xiao2021Meibo} & Images & $15$ & Nan & $2$ ophthalmologists \\
         Yeh C-H et al. ($2021$)~\cite{yeh2021Meibo} & Images & $398$ (train), $99$ (val) & $209$ & Trained clinician \\
         da Cruz LB et al. ($2020$)\cite{dacruz2020interferometer} & Tabular & $106$ ($10$-fold CV) & Nan & Optometrist  \\
         da Cruz LB et al. ($2020$)~\cite{dacruz2020ripleysk} & Tabular & $106$ ($10$-fold CV) & Nan & Optometrist  \\
         Fu P-I et al. ($2020$)~\cite{Fu2020LLT} & Tabular & $28$ & Nan & Nan (clinical test results, subjective report) \\
         Fujimoto K et al. ($2020$)~\cite{fujimoto2020comparison} & Tabular & $195$ & Nan & Nan (kerato-conjunctival staining for \ac{DED}) \\
         Maruoka S et al. ($2020$)~\cite{Maruoka2020Meibo} & Images & $221$ ($5$-fold CV) & Nan & $3$ eyelid specialists \\
         Prabhu SM et al.($2020$)~\cite{Prabhu2020Meibo} & Images & $600$ & $200$ & Clinical experts \\
         Stegmann H et al. ($2020$)~\cite{Stegmann2020TearMeniscus} & Images & $6658$ (images) ($5$-fold CV) & Nan & Experienced investigator \\ 
         Wei S et al. ($2020$)~\cite{wei2020therapeutic} & Images & $5000$* & $53$ ($3-5$ per patient) & Experienced investigator  \\
         Giannaccare G et al. ($2019$)~\cite{Giannaccare2019SubbasalNerve} & Tabular & Nan & $69$ & Experienced investigator~\cite{Chen2017ACCMed} \\
         Llorens-Quintana et al. ($2020$)~\cite{Llorens-QuintanaClara2019ANAA} & Images & $149$ & Nan & Clinicians \\
         Wang J et al. ($2019$)~\cite{Wang2019Meibo} & Images & $398$ (train) $99$ (val) & $209$ & Experienced clinician \\
         Yabusaki K et al. ($2019$)~\cite{yabusaki2019diagnose} & Tabular & $93$** & $45$** & Skilled ophthalmologist \\
         Yang J et al. ($2019$)~\cite{yang2019meniscus} & Images & $520$ & Nan & ImageJ software  \\
         Szyperski PD ($2018$)~\cite{Szyperski2018fractal} & Tabular & $110$ & Nan & Nan \\
         Hwang H et al. ($2017$)~\cite{hwang2017image} & Frames & $34$ & Nan & Meibomian gland expert \\
         Koprowski R et al. ($2017$)~\cite{Koprowski2017Meibo} & Images & $228$ (images) & Nan & Specialized clinicians \\
         Peteiro-Barral D et al. ($2017$)~\cite{peteiro2017evaluation} & Tabular & $105$ (LOO CV) & Nan & Experts  \\
         Koprowski R et al. ($2016$)~\cite{Koprowski20161Meibo} & Images & $172$ (images) & Nan & Ophthalmology expert \\
         Remeseiro B et al. ($2016$)~\cite{remeseiro2016iDEAS} & Tabular & Nan & $128$ & Optometrists \\
         Remeseiro B et al. ($2016$)~\cite{remeseiro2016CASDES} & Tabular & Sampled from test set & $50$ & $4$ optometrists \\
         Kanellopoulos AJ et al. ($2014$)~\cite{kanellopoulos2014invivo} & Tabular & $140$ & Nan & Ophthalmologist \\
         Ramos L et al. ($2014$)~\cite{ramos2014automatic} & Videos & $18$ & Nan & $2$/$4$ experts  \\
         Ramos L et al. ($2014$)~\cite{Ramos2014CCLRU} & Videos & $12$ & $6$ & $4$ experts  \\
         Remeseiro et al. ($2014$)~\cite{remeseiro2014tearfilm} & Tabular & $511$ ($10$-fold CV) & Nan & Experts \\
         García-Resúa C et al. ($2013$)~\cite{garcia2013new} & Tabular & $105$ ($6$-fold CV) & Nan & Experienced investigator  \\
         Rodriguez R et al. ($2013$)~\cite{Rodriguez2013Redness} & Tabular & $99$ (images) & Nan & $5$ trained investigators \\
         Koh YW et al. ($2012$)~\cite{koh2012detection} & Tabular & $28$*** & $27$*** & Experts  \\
         Yedidya T et al. ($2009$)~\cite{yedidya2009enforcing} & Videos & $22$ & Nan & Clinician  \\
         Yedidya T et al. ($2007$)~\cite{yedidya2007automatic} & Frames & $8$**** & Nan & Optometrist (evaluated $3$ of the $8$ patients)  \\
         Mathers WD et al. ($2004$)~\cite{mathers2004cluster} & Tabular & $513$ ($10$-fold CV) & Nan & Nan (clinical test results)  \\
         
         \hline
         \textbf{Biochemical Investigations} \\
         
         Cartes C et al. ($2019$)~\cite{cartes2019dry} & Tabular & $40$ (noise added) & $40$ (no noise) & Nan (clinical test results, subjective report)  \\
         Jung JH et al. ($2017$)~\cite{jung2017proteomic} & Tabular & $10$ & Nan & Ophthalmologist  \\
         Gonzalez N et al. ($2014$)~\cite{Gonzalez2014Protein} & Tabular & $70$\% of $93$** & $30$\% of $93$** & Nan (clinical tests) \\
         Grus FH et al. ($2005$)~\cite{grus2005seldi} & Tabular & $50$ \% of $159$ & $50$ \% of $159$ & Nan (clinical test results, subjective report)  \\
         Grus FH et al. ($1999$)~\cite{grus1999analysis} & Tabular & $30$ & $30$ & Nan (clinical test results, subjective report)  \\
         Grus FH et al. ($1998$)~\cite{grus1998clustering} & Tabular & $119$ & $\oplus$ & Nan (clinical test results, subjective report)  \\
         
         \hline
         \textbf{Demographical Investigations} \\
         
         Choi HR et al. ($2020$)~\cite{choi2020association} & Tabular & $2272$ & Nan & Nan (subjective report)  \\
         Nam SM et al. ($2020$)~\cite{nam2020explanatory} & Tabular & $80$ \% of $4391$ & $20$ \% of $4391$ & Ophtalmologist   \\
         Kaido M et al. ($2015$)~\cite{Kaido2015computer} & Tabular & $369$ & Nan & Dry eye specialists  \\
         
        \bottomrule
        
        \multicolumn{5}{p{.7\linewidth}}{Abbreviations: Nan = not available; val = validation; CV = crossvalidation; DED = dry eye disease; LOO = leave one out; * = pretraining images; ** = randomly selected samples, process repeated $10$ times; *** = randomly selected samples, process repeated $100$ times; **** = $3 - 5$ sequences of video per patient; $\oplus$ = For multivariate analysis model, but the number of samples was not mentioned}
    \end{tabular}
\end{table}
\end{landscape}
% -------------------------------
\subsection{Fluorescein tear break-up time}
% -------------------------------
Shorter break-up time indicates an unstable tear film and higher probability of \ac{DED}. Machine learning has been employed to detect dry areas in \ac{TBUT} videos and estimate \ac{TBUT}~\cite{yedidya2007automatic,yedidya2009enforcing,ramos2014automatic,Ramos2014CCLRU}. Use of the Levenberg-Marquardt algorithm to detect dry areas achieved an accuracy of $91$\% compared to assessments by an optometrist~\cite{yedidya2007automatic}. Application of Markov random fields to label pixels based on degree of dryness was used to estimate \ac{TBUT} resulting in an average difference of $2.34$ seconds compared to clinician assessments~\cite{yedidya2009enforcing}. Polynomial functions have also been used to determine dry areas, where threshold values were fine-tuned before estimation of \ac{TBUT}~\cite{ramos2014automatic}. This method resulted in more than $90$\% of the videos deviating by less than $\pm 2.5$ seconds compared to analyses done by four experts on videos not used for training~\cite{Ramos2014CCLRU}. Taken together, these studies indicate that \ac{TBUT} values obtained using automatic methods are within an acceptable range compared to experts. However, we only found four studies, all of them including a small number of subjects. Further studies are needed to verify the findings and to test models on external data.
% -------------------------------
\subsection{Interferometry and slit-lamp images}\label{sec:interferometry}
% -------------------------------
Interferometry is a useful tool that gives a snapshot of the status of the tear film lipid layer, which can be used to aid diagnosis of \ac{DED}.
Machine learning systems have been applied to interferometry and slit-lamp images for lipid layer classification based on morphological properties~\cite{garcia2013new,remeseiro2014tearfilm,remeseiro2016CASDES,remeseiro2016iDEAS,peteiro2017evaluation,dacruz2020interferometer,dacruz2020ripleysk}, estimation of the lipid layer thickness~\cite{hwang2017image,Fu2020LLT}, diagnosis of \ac{DED}~\cite{Szyperski2018fractal,yabusaki2019diagnose}, determination of ocular redness~\cite{Rodriguez2013Redness} and estimation of tear meniscus height~\cite{yang2019meniscus,DENG2021TearMeniscus}.

Diagnosis of \ac{DED} can be based on the following morphological properties: open meshwork, closed meshwork, wave, amorphous and color fringe~\cite{guillon1998lipid}. Most studies used these properties to automatically classify interferometer lipid layer images using machine learning. Garcia et al.\ used a K-nearest neighbors model trained to classify images resulting in an accuracy of $86.2$\%~\cite{garcia2013new}. Remeseiro et al.\ explored various \ac{SVM} models for use in final classification~\cite{remeseiro2014tearfilm,remeseiro2016CASDES,remeseiro2016iDEAS}.
In one of the studies, the same data was used for training and testing, which is not ideal~\cite{remeseiro2016CASDES}. Another study did not report the data their system was trained on~\cite{remeseiro2016iDEAS}.
Peteiro et al.\ evaluated images using five different machine learning models~\cite{peteiro2017evaluation}. In this study, the amorphous property was not included as one of possible classifications, as opposed to the other studies. A simple neural network achieved the overall best performance with an accuracy of $96$\%. However, because leave-one-out cross validation was applied, the model may have overfitted on the training data~\cite{hastie2009Unsupervised}. 
da Cruz et al.\ compared six different machine learning models and found that the random forest was the best classifier, regardless of the pre-processing steps used~\cite{dacruz2020interferometer,dacruz2020ripleysk}. The highest performance was achieved by application of Ripley's K function in the image pre-processing phase, and Greedy Stepwise technique used simultaneously with the machine learning models for feature selection~\cite{dacruz2020ripleysk}. Since all models were evaluated with cross validation, the system should be externally evaluated on new images before being considered for routine use in the clinic. 

Hwang et al.\ investigated whether tear film lipid layer thickness can be used to distinguish \ac{MGD} severity groups~\cite{hwang2017image}. Machine learning was used to estimate the thickness from Lipiscanner and slit-lamp videos with promising results. Images were pre-processed and the flood-fill algorithm and canny edge detection were applied to locate and extract the iris from the pupil. A significant difference between two \ac{MGD} severity groups was detected, suggesting that the technique could be used for the evaluation of \ac{MGD}. 
Keratograph images can also be used to determine tear film lipid layer thickness. Comparison of two different image analysis methods using a generalized linear model showed that there was a high correlation between the two techniques~\cite{Fu2020LLT}. The authors concluded that the simple technique was sufficient for evaluation of tear film lipid layer thickness. However, only $28$ subjects were included in the study. 

The use of fractal dimension estimation techniques was investigated for feature extraction from interferometer videos for diagnosis of \ac{DED}~\cite{Szyperski2018fractal}. The technique was found to be fast and had an \ac{AUC} value of $0.786$, compared to a value of $0.824$ for an established method (See~\Cref{sec:metrics_appendix,fig:auroc} for a description of the receiver operating characteristic curve).
Tear film lipid interferometer images were analysed using an \ac{SVM}~\cite{yabusaki2019diagnose}. Extracted features from the images were passed to the \ac{SVM} model, which classified the images as either healthy, aqueous-deficient \ac{DED}, or evaporative \ac{DED}. 
The agreement between the model and a trained ophthalmologist was high, with a reported Kappa value of $0.82$. The model performed best when detecting aqueous-deficient \ac{DED}.

Ocular redness is an important indicator of dry eyes. Only one of the reviewed studies described an automated system for evaluation of ocular redness associated with \ac{DED}~\cite{Rodriguez2013Redness}. Slit-lamp images were acquired from $26$ subjects with a history of \ac{DED}. Features representing the ocular redness intensity and horizontal vascular component were extracted with a Sobel operator. A multiple linear regression model was trained to predict ocular redness based on the extracted features. The system achieved an accuracy of $100$\%. The authors suggested that an objective system like this could replace subjective gradings by clinicians in multicentered clinical studies.

The tear meniscus contains $75 - 90$\% of the aqueous tear volume~\cite{Holly1985meniscus}. Consequently, the tear meniscus height can be used as a quantitative indicator for \ac{DED} caused by aqueous deficiency. When connected component labelling was applied to slit-lamp images, the Pearson's correlation between the predicted meniscus heights and an established software methodology (ImageJ~\cite{imageJref}) was high, ranging between $0.626$ and $0.847$~\cite{yang2019meniscus}. The machine learning system was found to be more accurate than four experienced ophthalmologists. The tear meniscus height can also be estimated from keratography images using a \ac{CNN}~\cite{DENG2021TearMeniscus}. The automatic machine learning system achieved an accuracy of $82.5$\% and was found to be more effective and consistent than a well-trained clinician working with limited time.

Many of the studies apply \ac{SVM} as their type of machine learning model without testing how other machine learning models perform. However, three of the studies tested several types of models and found that \ac{SVM} did not perform the best~\cite{peteiro2017evaluation,dacruz2020interferometer,dacruz2020ripleysk}. It is difficult to compare the studies due to different applications and evaluation metrics. 
Despite promising results, most of the studies~\cite{garcia2013new,remeseiro2014tearfilm,remeseiro2016CASDES,peteiro2017evaluation,dacruz2020interferometer,dacruz2020ripleysk,hwang2017image,Fu2020LLT,Szyperski2018fractal,Rodriguez2013Redness,yang2019meniscus} did not evaluate their systems on external data. The systems should be tested on independent data before they can be considered for clinical application. 
Moreover, some studies were small~\cite{Rodriguez2013Redness,Fu2020LLT} or pilots~\cite{yang2019meniscus,DENG2021TearMeniscus}, and the suggested models should be tested on a larger number of subjects. 
% -------------------------------
\subsection{In vivo confocal microscopy}\label{sec:ivcm}
% -------------------------------
\ac{IVCM} is a valuable non-invasive tool used to examine the corneal nerves and other features of the cornea~\cite{Cruzat2016IVCM}.
\ac{IVCM} images were used in a small study to assess characteristics of the corneal subbasal nerve plexus for diagnosis of \ac{DED}~\cite{Giannaccare2019SubbasalNerve}. Application of random forest and a deep neural network~\cite{Chen2017ACCMed} gave promising results with an \ac{AUC} value of $0.828$ for detecting \ac{DED}~\cite{Giannaccare2019SubbasalNerve}.
\ac{IVCM} images of corneal nerves can also be analyzed by machine learning models to estimate the length of the nerve fiber~\cite{wei2020therapeutic}. Authors used a \ac{CNN} with a U-net architecture that had been pre-trained on more than $5,000$ \ac{IVCM} images of corneal nerves. The model showed that nerve fiber length was significantly longer after intense pulsed light treatment in \ac{MGD} patients, which agreed with manual annotations from an experienced investigator with an \ac{AUC} value of $0.96$ and a sensitivity of $0.96$. 
High-resolution \ac{IVCM} images were also used to detect obstructive \ac{MGD}~\cite{Maruoka2020Meibo}. Combinations of nine different \acp{CNN} were trained and tested on the images using $5$-fold cross validation. Classification by the models was compared to diagnosis made by three eyelid specialists. The best performance was achieved when four different models were combined, with high sensitivity, specificity and \ac{AUC} values, see~\Cref{tab:clinicaltable_1}. These promising results suggest that \acp{CNN} can be useful for detection and evaluation of \ac{MGD}. 
Deep learning methods such as \acp{CNN} have the advantage that feature extraction from the images prior to analysis is not required as this is performed automatically by the model. 

\ac{IVCM} images have been investigated for changes in immune cells across different severities of \ac{DED} for diagnostic purposes~\cite{aggarwal2021immunecell}. A generalized linear model showed significant differences in dendritic cell density and morphology between \ac{DED} patients and healthy individuals, but not between the different \ac{DED} subgroups, see~\Cref{tab:clinicaltable_1}. While results using machine learning to interpret \ac{IVCM} images are promising, larger clinical studies are needed to validate findings before clinical use can be considered. 
% -------------------------------
\subsection{Meibography}
% -------------------------------
The meibomian glands are responsible for producing meibum, important for protecting the tear fluid from evaporation. Reduced secretion of meibum due to a reduced number of functional meibomian glands and/or obstruction of the ducts is a major cause of evaporative \ac{DED} and \ac{MGD}. Classification of meibomian glands using meibography is routine for experienced experts, but this is not the case for all clinicians. Moreover, automatic methods can be faster than human assessment. 

Meibography images may require several pre-processing steps before they can be classified. One study trained an \ac{SVM} on extracted features from the images~\cite{koh2012detection}. Pre-processing included the dilation, flood-fill, skeletonization and pruning algorithms. The model achieved a sensitivity of $0.979$ and specificity of $0.961$. However, in contrast to all other image analysis methods, this method is not completely automatic as the images need to be manipulated manually before they are passed on to the system. 

A combination of Otsu's method and the skeletonization and watershed algorithms was useful in automatically quantifying meibomian glands~\cite{Koprowski20161Meibo}. This method was faster than an ophthalmologist and achieved a sensitivity and specificity of $0.993$ and $0.975$, respectively. Another automatic method applied B{\'e}zier curve fitting as part of the analysis~\cite{Koprowski2017Meibo}. The reported sensitivity was $1.0$, while the specificity was $0.98$. Xiao et al.\ sequentially applied a Prewitt operator, Graham scan, fragmentation and skeletonization algorithms for image analysis to quantify meibomian glands~\cite{xiao2021Meibo}. The agreement between the model results and two ophthalmologists was high with Kappa values larger than $0.8$ and low false positive rates ($<0.06$). The false negative rate was $0.19$, suggesting that some glands were missed by the method. A considerable weakness of this study was that only $15$ images were used for model development, and consequently it might not work well on unseen data. 
Another study automatically graded \ac{MGD} severity using a Sobel operator, polynomial functions, fragmentation algorithm and Otsu's method~\cite{Llorens-QuintanaClara2019ANAA}. While the method was found to be faster, the results were significantly different from clinician assessments. 

Deep learning approaches were used by four studies evaluating meibomian gland features~\cite{Wang2019Meibo,Prabhu2020Meibo,yeh2021Meibo,Khan2021image}. These systems are fully automated and apply some of the latest technologies within image analysis. Wang et al.\ used four different \acp{CNN} to determine meibomian gland atrophy~\cite{Wang2019Meibo}. The \acp{CNN} were trained to identify meibomian gland drop-out areas and estimate the percentage atrophy in a set of images. Comparison of model predictions with experienced clinicians indicated that the best \ac{CNN} (ResNet50 architecture) was superior. Yeh et al.\ developed a method to evaluate meibomian gland atrophy by extracting features from meibography images with a special type of unsupervised \ac{CNN} before application of a K-nearest neighbors model to allocate a meiboscore~\cite{yeh2021Meibo}. The system achieved an accuracy of $80.9$\%, outperforming annotations by the clinical team. Moreover, hierarchical clustering of the extracted features from the \ac{CNN} could show relationships between meibography images. Another study used a \ac{CNN} to automatically assess meibomian gland characteristics~\cite{Prabhu2020Meibo}. Images from two different devices collected from various hospitals were used to train and evaluate the \ac{CNN}. This is an example of uncommonly good practice, as most medical \ac{AI} systems are developed and evaluated on data from only one device and/or hospital.
The only study to use a \ac{GAN} architecture tested it on infrared 3D images of meibomian glands in order to evaluate \ac{MGD}~\cite{Khan2021image}. Comparing the model output with true labels, the performance scores were better than for state of the art segmentation methods. The Pearson correlations between the new automated method and two clinicians were $0.962$ and $0.968$.

Four of the studies did not evaluate their proposed systems on external data~\cite{Koprowski20161Meibo, Koprowski2017Meibo,Llorens-QuintanaClara2019ANAA, xiao2021Meibo}. Since the number of images used for model development was limited, the models can have overfit, and external evaluations should be performed to test how well the systems generalize to new data.
% -------------------------------
\subsection{Tear osmolarity}
% -------------------------------
Tear osmolarity is a measure of tear concentration, and high values can indicate dry eyes. Cartes et al.~\cite{cartes2019dry} investigated use of machine learning to detect \ac{DED} based on this test. Four different machine learning models were compared. Noise was added to osmolarity measurements during the training phase, while original data without noise was used for final evaluation. The logistic regression model achieved $85$\% accuracy. However, since the models were trained and tested on the same data, the reported score is most likely not representative for how well the model generalizes to new data.  
% -------------------------------
\subsection{Proteomic analysis}
% ------------------------------- 
Proteomic analysis describes the qualitative and quantitative composition of proteins present in a sample.
Grus et al.\ compared tear proteins in individuals with diabetic \ac{DED}, non-diabetic \ac{DED} and healthy controls for discrimination between the groups~\cite{grus1998clustering}. The authors used discriminant analysis and principal component analysis combined with k-means clustering. Both models achieved low accuracies when predicting all three categories. However, classification into \ac{DED} and non-\ac{DED} achieved accuracies of $72$\% and $71$\% for discriminant analysis and k-means clustering, respectively. 
In another study by the same group, tear proteins analyzed using deep learning discriminated subjects as healthy or having \ac{DED} with an accuracy of $89$\%~\cite{grus1999analysis}. An accuracy of $71$\% was achieved using discriminant analysis. A combination of discriminant analysis for detecting the most important proteins and a deep neural network for classification was also investigated~\cite{grus2005seldi}. High accuracy, sensitivity and specificity were reported.
Discriminant analysis was also used by Gonzalez et al.\ in analysis of the tear proteome~\cite{Gonzalez2014Protein}. The most important proteins were selected to train an artificial neural network to classify tear samples as aqueous-deficient \ac{DED}, \ac{MGD} or healthy. The model gave an overall accuracy of $89.3$\%. Principal component analysis yielded good separation of healthy controls, aqueous-deficient \ac{DED} and \ac{MGD} data-points, indicating that the proteins were good candidates for classification of the three conditions. This system achieved the highest accuracy of all the reviewed proteomic studies. Considered together, the results from the four studies~\cite{grus1998clustering, grus1999analysis, grus2005seldi,Gonzalez2014Protein} suggest that neural networks applied alone or together with other techniques perform better than discriminant analysis for detecting \ac{DED}-related protein patterns in the tear proteome.

Jung et al.\ used a network model based on modularity analysis to describe the tear proteome with respect to immunological and inflammatory responses related to \ac{DED}~\cite{jung2017proteomic}. In this study, patterns in tears and lacrimal fluid were investigated in patients with \ac{DED}. 
Since only $10$ subjects were included, the study should be performed on a larger cohort of patients to verify the results.
% ------------------------------- 
\subsection{Optical coherence tomography}
% ------------------------------- 
Thickening of the corneal epithelium can be a sign of abnormalities in the cornea. Moreover, corneal thickness could potentially be a marker for \ac{DED}. 
Kanellopoulos et al.\ developed a linear regression model to look for possible correlations between corneal thickness metrics measured using \ac{AS-OCT} and \ac{DED}~\cite{kanellopoulos2014invivo}. However, neither the model predictions nor performance were reported, making it difficult to assess the usefulness of the study. 
The type of instrument used to determine the corneal thickness was found to affect the results~\cite{fujimoto2020comparison}. Measurements from \ac{AS-OCT} and Pentacam were compared and  multivariable regression was used to detect differences between the two techniques regarding the measured central corneal thickness and the thinnest corneal thickness. Individuals with mild \ac{DED}, severe \ac{DED} and healthy subjects were examined. The two techniques gave significantly different results in terms of the resulting $\beta$-coefficients in the multivariable regression model for individuals with severe \ac{DED}.
Images from clinical examinations with \ac{AS-OCT} were used to diagnose \ac{DED}~\cite{ELSAWY2021252}. A pretrained VGG19 \ac{CNN}~\cite{vgg2017Elsawy} was fine-tuned using separate images for training and validation. 
Two similar \ac{CNN} models were developed, and evaluation was performed on an external test set. Both achieved impressively high performance scores. The \ac{AUC} values were $0.99$ and $0.98$.  
This is one out of two studies in this review that used an independent test sets after model development. Such practice is essential for a realistic impression of how well the model generalizes to new data not used during model development. 
The good performance is likely linked to the large amounts of training data ($29,000$ images), which is essential for deep learning methods. Most of the reviewed studies use significantly smaller data sets, which constitutes a disadvantage.  
Stegmann et al.\ analysed \ac{OCT} images from healthy subjects for automatic detection of the lower tear meniscus~\cite{Stegmann2020TearMeniscus}. Two different \acp{CNN} were trained and evaluated using $5$-fold cross validation. The tear menisci detected by the models were compared to evaluations from an experienced grader. The best \ac{CNN} achieved an average accuracy of $99.95$\%, sensitivity of $0.9636$ and specificity of $0.9998$. The system is promising regarding fast and accurate segmentation of \ac{OCT} images. However, more images from different \ac{OCT} systems, including non-healthy subjects, should be used to verify and improve the analysis.

The two studies~\cite{vgg2017Elsawy,Stegmann2020TearMeniscus} showed that \acp{CNN} could be an appropriate tool for image analysis. \acp{CNN} are likely to increase in popularity within the field of \ac{DED} due to promising results for solving image related tasks, including feature extraction.
% ------------------------------- 
\subsection{Other clinical tests} 
Machine learning models were used to analyse results from a variety of clinical tests to expand understanding of the \ac{DED} process~\cite{mathers2004cluster}. The study included subjects with \ac{DED} and healthy subjects. Subjective cutoff values from clinical tests were used to assign subjects to the \ac{DED} class. Hierarchical clustering and a decision tree were applied sequentially to group the subjects based on their clinical test results. The resulting groups were compared to the original groups. Because the analysis was based on objective measurements, it could be used to develop more objective diagnostic criteria. This could lead to earlier detection and more effective treatment of \ac{DED}. 
% ------------------------------- 
\subsection{Population surveys}
% ------------------------------- 
Population surveys can provide valuable insight regarding the prevalence of \ac{DED} and help detect risk factors for developing the disease. Japanese visual terminal display workers were surveyed with the objective of detecting \ac{DED}~\cite{Kaido2015computer}. Dry eye exam data and subjective reports were used for diagnosis. This was passed to a discriminant analysis model. When compared to diagnosis by a dry eye specialist, the model showed a high sensitivity of $0.931$, but low specificity of $0.437$. This is a very low specificity, but is not necessarily bad if the aim is to detect as many cases of \ac{DED} as possible and there is less concern about misclassification of healthy individuals. 
Data from a national health survey were analysed in order to detect risk factors for \ac{DED}~\cite{nam2020explanatory}. Here, individuals were regarded as having \ac{DED} if they had been diagnosed by an ophthalmologist, and were experiencing dryness. Feature modifications were performed by a decision tree, and the most important features were selected using lasso. $\beta$-coefficients from a logistic regression trained on the most important features were used to rank the features. Women, individuals who had received refractive surgery and those with depression were detected as having the highest risk for developing \ac{DED}. Even though the models in the study were trained on data from more than $3500$ participants, the reported performance scores were among the poorest in this review with a sensitivity of $0.66$ and a specificity of $0.68$. A possible reason could be that the selected features were not ideal for detecting \ac{DED}. However, the detected risk factors have previously been shown to be associated with \ac{DED}~\cite{matossian2019dry,dartt2004dysfunctional,wan2016depression}. The findings suggest that the data quality from population surveys might not be as high as in other types of studies, which could lead to misinterpretation by the machine learning model.   

The association between \ac{DED} and dyslipidemia was investigated by combining data from two population surveys in Korea in~\cite{choi2020association}. A generalized linear model was used to investigate linear characteristics between features and the severity of \ac{DED}. The model showed significant increase in age, blood pressure and prevalence of hypercholesterolemia over the range from no \ac{DED} to severe \ac{DED}. Evaluation of the association between dyslipidemia and \ac{DED} using linear regression showed that the odds ratio for men with dyslipidemia was higher than $1$ compared to men without dyslipidemia. This association was not found in women. The study results suggest a positive association between \ac{DED} and dyslipidemia in men, but not in women. 
% ---------------------------------------------
\subsection{Future perspectives}
In order to benchmark existing and future models, we advocate that the field of \ac{DED} should have a common, centralized and openly available data set for testing and evaluation. The data should be fully representative for the relevant clinical tests. In order to ensure that models are applicable to all populations of patients, medical institutions, and types of equipment around the world, they must be evaluated on data from different demographic groups of patients across several clinics and, if relevant, from different medical devices. Moreover, the test data set should not be available for model development, but only for final evaluation. 
A common standard on these processes will increase the reproducibility and comparability of studies. In addition, a cross hospitals/centers data set would solve important challenges of applying \ac{AI} in clinical practice, such as metrics not reflecting clinical applicability, difficulties in comparing algorithms, and underspecification. These have all been identified as being among the main obstacles for adoption of any medical \ac{AI} system in clinical practice~\cite{d2020underspecification,kelly2019key}.

A possible challenge regarding implementation in the clinic is that hospitals do not necessarily use the same data platforms, which might prevent widespread use of machine learning systems. Consequently, solutions for implementing digital applications across hospitals should be considered. 

Model explanations are important in order to understand why a complex machine learning model produces a certain prediction. For healthcare providers to trust the systems and decide to use them in the clinic, the systems should provide understandable and sound explanations of the decision-making process. Moreover, they could assist clinicians when making medical decisions~\cite{lundberg2018hypoxemia}. When developing new machine learning systems within \ac{DED}, effort should be made to present the workings of the resulting models and their predictions in an easy to interpret fashion. 
% ---------------------------------------------
\section{Conclusions}\label{sec:concl}
We observed a large variation in the type of clinical tests and the type of data used in the reviewed studies. This is also true regarding the extent of pre-processing applied to the data before passing it to the machine learning models. The studies analysing images can be divided into those applying deep learning techniques directly on the images, and those performing extensive pre-processing and feature extraction before the data is passed to the machine learning model in a tabular format. The number of studies belonging to the first group has increased significantly over the past $3$ years. As deep learning techniques become more established, these will probably replace more traditional image pre-processing and feature extraction techniques. 

We noted that there was a lack of consensus regarding how best to perform model development, including evaluation. This made it difficult to estimate how well some models will perform in the clinic and with new patients, and also to compare the different models. Comparison was further complicated by the use of different types of performance scores. In addition there was no culture of data and code sharing, which makes reproducibility of the results impossible. For the future, focus should be put on establishing data and code sharing as a standard procedure.

In conclusion, the results from the different studies' machine learning models are promising, although much work is still needed on model development, clinical testing and standardisation. \ac{AI} has a high potential for use in many different applications related to \ac{DED}, including automatic detection and classification of \ac{DED}, investigation of the etiology and risk factors for \ac{DED}, and in the detection of potential biomarkers. Effort should be made to create common guidelines for the model development process, especially regarding model evaluation. Prospective testing is recommended in order to evaluate whether proposed models can improve the diagnostics of \ac{DED}, and the health and quality of life of patients with \ac{DED}.

\section*{Disclosure}
The authors report no conflicts of interest. 

\bibliography{bibliography}
% ---------------------------------------------
\appendix
\begin{appendices}
\numberwithin{equation}{section}
% ---------------------------------------------
\section{Supporting information}
% ---------------------------------------------

% ---------------------------------------------
\subsection{Performance scores used}\label{sec:metrics_appendix}
\begin{figure}[!tb]
    \captionsetup[subfigure]{justification=centering}
    \begin{subfigure}{.49\textwidth}{
        \centering
        \includegraphics[height=5cm]{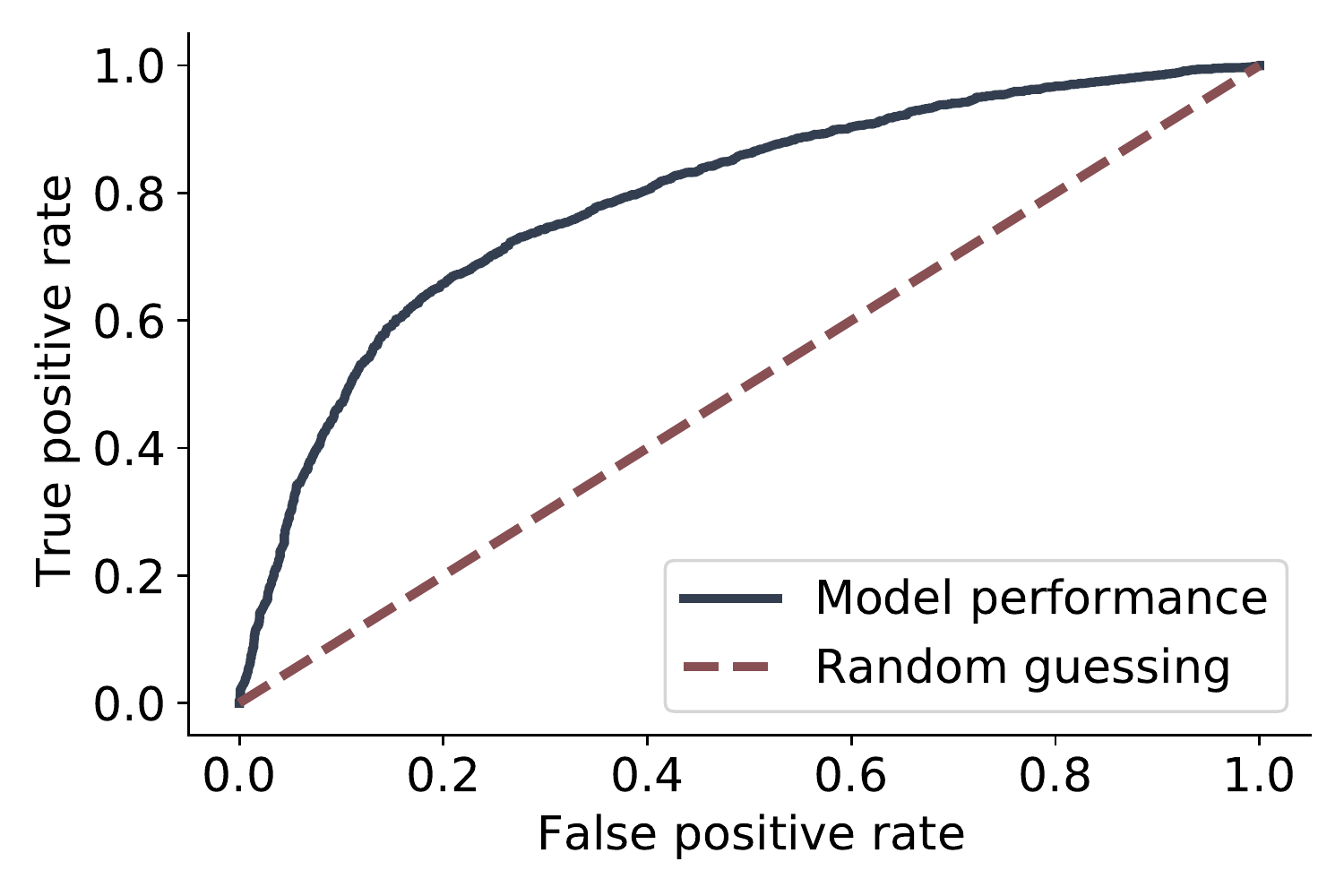}
        \caption{\label{fig:auroc}}}
    \end{subfigure}
    \begin{subfigure}{.49\textwidth}{
        \includegraphics[height=5cm]{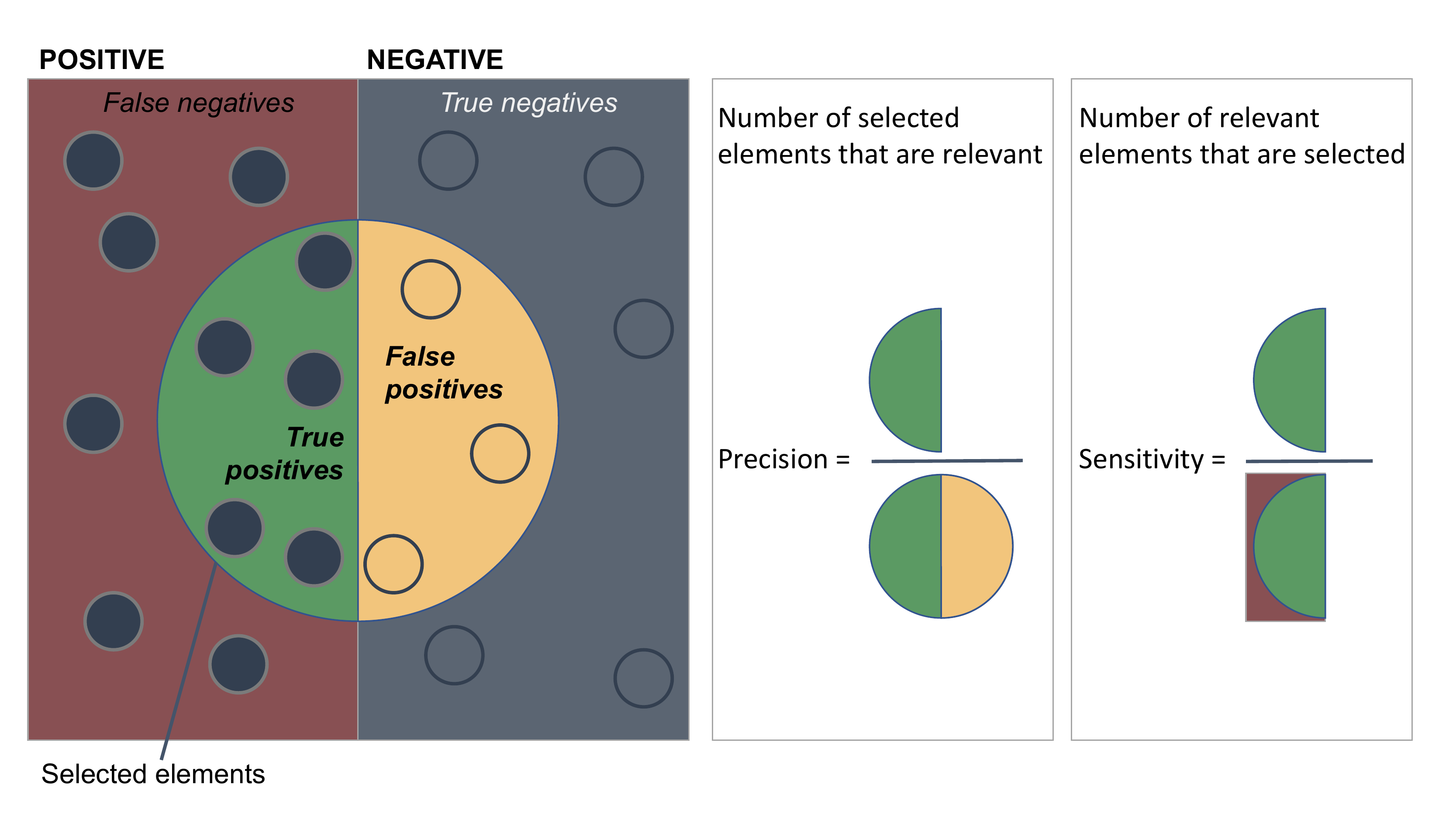}
     \caption{\label{fig:prec_rec}}}
    \end{subfigure}
    \caption{(\protect\subref{fig:auroc}) 
    A receiver operating characteristic curve, and
    (\protect\subref{fig:prec_rec})
    A visual representation of the sensitivity, eq.~\eqref{eq:recall}, and the precision, eq.~\eqref{eq:precision}, highlighting the trade-off between the two.}
\end{figure}

% ---------------------------------------------
If there are two categories available, the task is referred to as binary classification, while more than two categories is referred to as multi-class. For binary classification, the true outcome belongs to one of two categories, e.g., healthy or ill, often referred to as positive (P) or negative (N). A binary classifier assigns new data instances to these two categories, and the prediction can be either true (T), meaning correct, or false (F), meaning incorrect. The outcome can then belong to one of the four categories \ac{TP}, \ac{TN}, \ac{FP} and \ac{FN}, and sum to the total number of instances in the data set.
From these, we can calculate a variety of performance scores, some of which are listed in~\Cref{sec:metrics}. We provide mathematical expression for these below. The remaining performance scores encountered in the reviewed studies are outlined after. 
\begin{eqnarray}
    \label{eq:ppv}
    \text{Positive predictive value} &=& \frac{\ac{TP}}{\ac{TP} + \ac{FP}}\\
    \label{eq:npv}
    \text{Negative predictive value} &=& \frac{\ac{TN}}{\ac{TN} + \ac{FN}}\\
    \label{eq:accuracy}
    \text{Accuracy} &=& \frac{\ac{TP} + \ac{TN}}{\ac{TP} + \ac{FP} + \ac{TN} + \ac{FN}} \\
    \label{eq:recall}
    \text{Sensitivity} &=& \frac{\ac{TP}}{\ac{TP} + \ac{FN}}\\
    \label{eq:precision}
    \text{Precision} &=& \frac{\ac{TP}}{\ac{TP} + \ac{FP}}\\
    \label{eq:specificity}
    \text{Specificity} &=& \frac{\ac{TN}}{\ac{TN} + \ac{FP}}\\
    \label{eq:f1_score}
    \text{F1 score} &=& \frac{2 \times \ac{TP}}{2 \times \ac{TP} + \ac{FP} + \ac{FN}}\\
    \label{eq:fpr}
    \text{False positive rate} &=& \frac{\ac{FP}}{\ac{FP} + \ac{TN}}\\
    \label{eq:fnr}
    \text{False negative rate} &=& \frac{\ac{FN}}{\ac{FN} + \ac{TP}}\,.
\end{eqnarray}
Although binary classification tasks involve assigning instances to one of two classes, e.g., $0$ and $1$, most machine learning classifiers can output the distance of an instance to the decision boundary, i.e., a decimal number in the interval $[0,1]$. A common interpretation of this number is class probability or classification confidence, meaning that an output close to either number indicates a confident classification, while an output closer to the classification threshold indicates that the classifier is not capable of assigning the instance to a class. The classification threshold is thus the numerical value that separates the two classes, and the confusion matrix entries vary with this threshold. Unless otherwise specified, its value is usually $0.5$. Here, we introduce two metrics that can be constructed by varying this threshold from $0$ to $1$. 
First, the receiver operating characteristic curve is constructed from the curves of the true and false positive rates obtained by varying the classification threshold. Optimally, the true positive rate is $1$ for any threshold, while a classifier which always guesses randomly produces a diagonal line, as shown in~\Cref{fig:auroc}. The \ac{AUC} value is calculated by summing the area under the receiver operating characteristic curve, and its maximum value is $1$.

There is a trade-off between the precision and sensitivity: A high precision minimizes the false positives, which might result in missing positive instances, while a high sensitivity minimizes the false negatives, which can result in a increased number of false alarms. Which one should be prioritised depends on the problem at hand, and a study prioritising or reporting only one of these should argue why. The precision and the sensitivity are visualised in~\Cref{fig:prec_rec}, which highlights the trade-off between the two. They can be combined into a single number, by plotting them against each other for different classification threshold values and calculating the area under the resulting so-called precision-recall curve. 

\smallskip
\textit{Pearson's correlation coefficient} measures the linear correlation between two data sets, and is calculated as
\begin{equation}
    \text{r} = \frac{\Sigma (x_i - \overline{x}) \times (y_i - \overline{y})}{\sqrt{\Sigma (x_i - \overline{x})^2 \times (y_i - \overline{y})^2}} \, ,
\end{equation}
where $r$ is the Pearson's correlation coefficient, $x_i$ and $y_i$ are the observed values in each data set and $\overline{x}$ and $\overline{y}$ are the mean values for each data set. 
The value ranges from $-1$ to $1$, where $-1$ indicates perfect negative linear correlation and $1$ perfect positive linear correlation, while $0$ indicates no linear correlation between the data.
For binary classification, Pearson's correlation coefficient takes on a simple form, referred to as Matthews correlation coefficient~\cite{mcc}. It measures the correlation between the true and predicted class instances, and ranges from $-1$ to $1$. Here, $0$ indicates that the classifier guesses randomly, and $1$ and $-1$ indicate complete agreement and disagreement, respectively, between the model predictions and the true outcome. It can be calculated from the confusion matrix entries as
\begin{equation}
    \text{Matthews correlation coefficient} = \frac{\ac{TP} \times \ac{TN} - \ac{FP} \times \ac{FN}}{\sqrt{(\ac{TP} + \ac{FP})(\ac{TP} + \ac{FN})(\ac{TN}+\ac{FP})(\ac{TN}+\ac{FN})}} \,.
\end{equation}
\smallskip
The \textit{concordance correlation coefficient} measures the agreement between two data sets by measuring the variation around the $45$ degrees concordance line through the origin~\cite{Lin1989ConcordanceCorr}. The value ranges between $1$ and $-1$. When the two data sets share mean and standard deviation, the concordance correlation coefficient equals the Pearsons's correlation coefficient. In all other cases, the concordance correlation coefficient will be lower than the Pearson's correlation coefficient. The value is calculated as
\begin{equation}
    \text{Concordance correlation coefficient} = \frac{2s_{xy}}{{s_x}^2 + {s_y}^2 + (\overline{x} - \overline{y})^2}\, ,
\end{equation}
where $\overline{x}$ and $\overline{y}$ are the mean values of the two data sets $x$ and $y$, ${s_x}^2$ and ${s_y}^2$ are the variances for each data set and ${s_{xy}}^2$ is the covariance between the data sets~\cite{Lin1989ConcordanceCorr}. 

\smallskip
\textit{Root mean squared error} is commonly used for regression problems and represents the difference between the model predictions and the observed values. The value is calculated as
\begin{equation}
    \text{Root mean squared error} = \sum\limits_{i=1}^n \frac{(\hat{y}_i - y_i)^2}{n}\, ,
\end{equation}
where $n$ is the number of instances in the data set and $\hat{y}_i$ and $y_i$ is the model prediction and observed value for instance $i$, respectively. 

\begin{figure}[tb!]
    \centering
    \includegraphics[width=\textwidth]{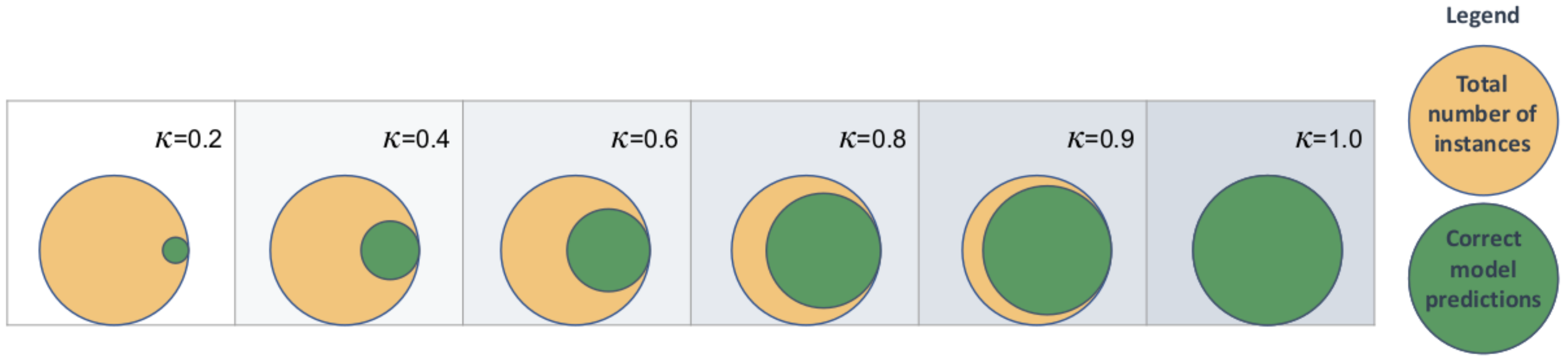}
    \caption{\label{fig:kappa}
    Kappa values for different degrees of agreement. The illustration is based on~\citep[Figure 2]{kappaIllustration}.}
\end{figure}

The \textit{Kappa index} measures the agreement between two raters, e.g., the model predictions and labels during classification~\cite{kappa_index}. It is calculated as
\begin{equation}\label{eq:kappa}
    \kappa = \frac{p_o - p_e}{1 - p_e} \,,
\end{equation}
where $p_o$ is the observed probability of agreement, which equals the accuracy defined in eq.~\eqref{eq:accuracy}, and $p_e$ is the expected probability of agreement due to chance, defined as 
\begin{equation}\label{eq:p_e}
    p_e = \frac{(\ac{TP} + \ac{FP}) \times (\ac{TN} + \ac{FN}) \times (\ac{FN} + \ac{TP}) \times (\ac{FP} + \ac{TP})}{\text{Total} \times \text{Total}} \,.
\end{equation} 
where $Total$ is the total number of instances.
The highest possible value is $1$, representing perfect agreement, and values above $0.8$ are typically regarded as excellent~\cite{kappa_index}. An illustration of the $\kappa$ index values for the proportion of correct model predictions is provided in~\Cref{fig:kappa}.

\textit{Cram{\'e}r's V} measures the association between two categorical variables that belong to more than two categories each. When there are two categories for each variable, Cram{\'e}r's V equals the $\varphi$ coefficient~\cite{Akoglu2018CramerGuide}. It is calculated via
\begin{equation}
    \text{Cram{\'e}r's V} = \sqrt{\frac{\frac{{\chi}^2}{n}}{\text{Min(cat1 - 1, cat2 -1)}}} \,,
\end{equation}
where ${\chi}^2$ is the usual chi-squared statistic, $n$ is the number of instances, and $cat1$ and $cat2$ are the number of possible categories for each variable.
The value ranges from $0$ to $1$, representing no and perfect correlation between the variables, respectively~\cite{CramerV1945}. 

In hypothesis testing, the $p$-value is the probability under a specific model of obtaining test results at least as extreme as those observed, under the assumption that the null hypothesis $H_0$ is true. $H_0$ is commonly defined as no difference between two data sets, while the alternate hypothesis $H_a$ states that there is a difference. Consequently, a low $p$-value  indicates that the result is not likely under the null hypothesis, and thus strengthens the belief in $H_a$~\cite{introStatistics}.

\begin{figure}[t]
    \centering
    \includegraphics[width=\textwidth]{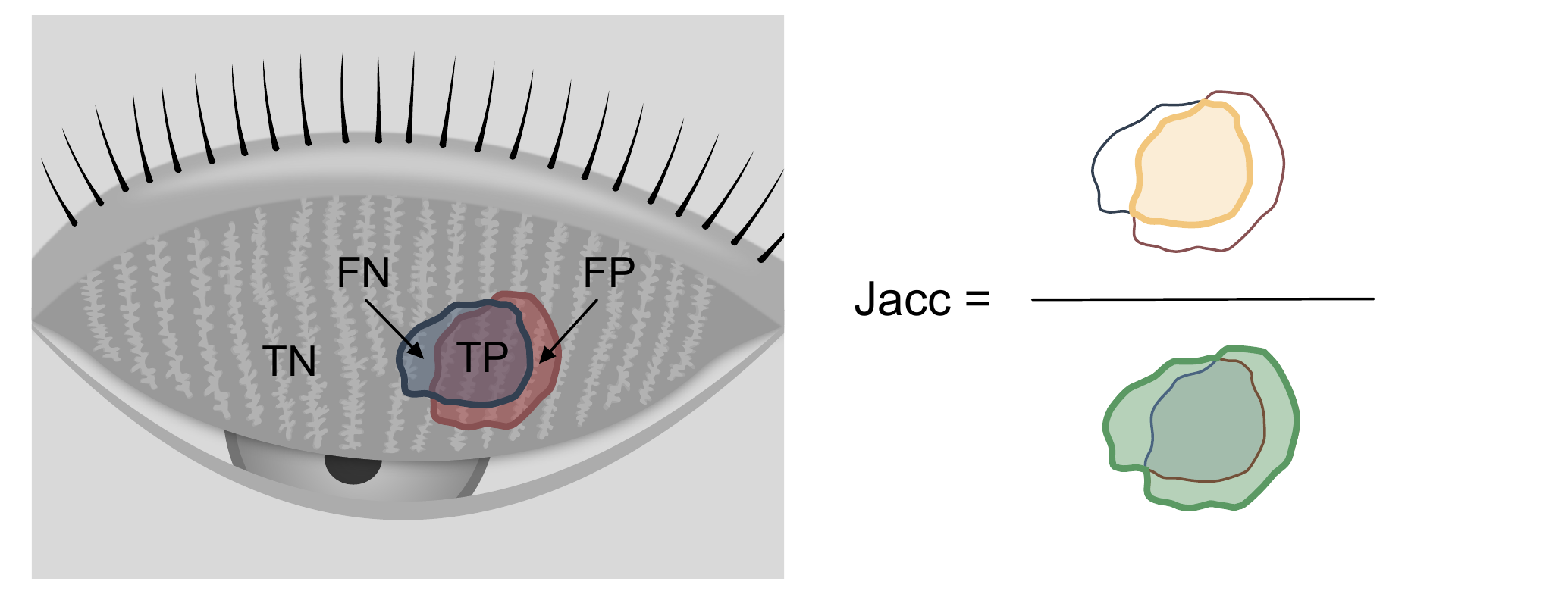}
    \caption{\label{fig:jaccard} A visual representation of the Jaccard index. FN = False Negative; TN = True Negative; TP = True Positive; FP = False Positive; Jacc = Jaccard index. 
    }
\end{figure}

The \textit{Average Pompeiu-Hausdorff distance} reflects the distance between estimated values and true values in a metric space~\cite{pomp_hausd}. Lower values imply small differences between the two metric spaces. The Pompieu-Hausdorff distance $H$ between the subsets $a$ and $b$ is calculated via
\begin{equation}\label{eq:aph}
    H(a,b) = max(H(a,b), H(b,a)) \,.
\end{equation}
The aggregated \textit{Jaccard index} is an extension of the global Jaccard index also used to measure the similarities between two sample sets~\cite{jaccardOrig1912}. A high value indicates small differences between the sample sets. The calculation of the aggregated Jaccard index is described by Kumar et al.~\cite{agg_jaccard}, and~\Cref{fig:jaccard} shows a visualisation. 
For image segmentation, the \textit{support} for a segmented area can be calculated as the number of pixels in the segmented area divided by the number of background pixels~\cite{Stegmann2020TearMeniscus}. 

\subsection{Measuring model uncertainty}
Uncertainty estimates are useful in order to evaluate how certain a machine learning model is about the predictions. High uncertainty might suggest that a human expert also should have a look at the instance~\cite{uncertainty_ml_med}. Among the reviewed studies, some choose not to use the model predictions of \ac{DED} when the predicted probabilities are too close to $0.5$, reflecting that the model is uncertain~\cite{Kaido2015computer}. Others report the standard deviation of the model performance scores~\cite{yedidya2007automatic, dacruz2020interferometer, dacruz2020ripleysk, koh2012detection, xiao2021Meibo, yeh2021Meibo,Stegmann2020TearMeniscus, yabusaki2019diagnose,Rodriguez2013Redness}. Some computes the confidence intervals for the model performance scores~\cite{ELSAWY2021252, fujimoto2020comparison, nam2020explanatory, Rodriguez2013Redness,Rodriguez2013Redness}. A comprehensive discussion about quantifying uncertainty for medical machine learning models can be found in~\cite{uncertainty_ml_med}.

\end{appendices}
\end{document}